\documentclass[journal]{IEEEtran}

\IEEEoverridecommandlockouts                              
{Fang \MakeLowercase{\textit{et al.}}: Efficient Jacobian-Based Inverse Kinematics of Soft Robots by Learning} 
\usepackage{algorithm}
\usepackage{float}
\usepackage{multicol}
\usepackage{multirow}
\usepackage{algpseudocode}
\usepackage{color}
\usepackage{xcolor}
\usepackage{graphicx}
\usepackage{bm}
\usepackage{amsmath}
\usepackage{amssymb}
\usepackage{tabularx}
\usepackage{subcaption}
\usepackage{url}
\usepackage{cite}
\usepackage{prettyref}
\usepackage{booktabs}
\usepackage{siunitx}
\usepackage{makecell}
\usepackage{comment}

\newrefformat{fig}{Figure~\ref{#1}}
\newrefformat{sec}{Section~\ref{#1}}
\newrefformat{tab}{Table~\ref{#1}}
\newrefformat{alg}{Algorithm~\ref{#1}}
\newrefformat{eqn}{Equation~\ref{#1}}

\DeclareMathOperator*{\argmin}{\arg\!\min}

\usepackage[normalem]{ulem}
\newcommand{\rev}[2]{\textcolor{red}{\sout{#1}}\textcolor{blue}{#2}}

\usepackage[draft]{hyperref}
\hypersetup{
    colorlinks,
    linkcolor={blue!80!black},
    citecolor={blue!80!black},
    urlcolor={blue!100!black}
}

\makeatletter
\newcommand{\printfnsymbol}[1]{%
  \textsuperscript{\@fnsymbol{#1}}%
}
\makeatother

\title{Efficient Jacobian-Based Inverse Kinematics with Sim-to-Real Transfer of Soft Robots by Learning}
\author{Guoxin Fang,~\IEEEmembership{Member,~IEEE}, 
Yingjun Tian, Zhi-Xin Yang,~\IEEEmembership{Member,~IEEE}, \\
Jo M.P. Geraedts, 
and Charlie C. L. Wang,~\IEEEmembership{Senior Member,~IEEE}
\thanks{Manuscript received May 20, 2021; revised Febuary 27, 2022; accepted May 09, 2022. The project is partially supported by the startup fund of the Digital Manufacturing Lab provided by the University of Manchester. \textit{(corresponding author: Charlie C.L. Wang)}}
\thanks{Guoxin Fang and Jo M. P. Geraedts are with the Faculty of Industrial Design Engineering, Delft
University of Technology, 2628 CE Delft, The Netherlands. (e-mail: G.Fang1@tudelft.nl; j.m.p.geraedts@tudelft.nl).}
\thanks{Yingjun Tian and Charlie C. L. Wang are with the Department
of Mechanical, Aerospace and Civil Engineering, The University
of Manchester, M13 9PL Manchester, U.K. (e-mail: yingjun.tian@postgrad.manchester.ac.uk; changling.wang@manchester.ac.uk).}
\thanks{Zhi-Xin Yang is with the State Key Laboratory of Internet of Things for Smart City and the Deptartment of Electromechanical Engineering, University of Macau, Avenida da Universidade Taipa, Macau, China. (e-mail: zxyang@um.edu.mo).}
\thanks{This article has supplementary video provided by the authors, available online at https://doi.org/10.1109/TMECH.2022.3178303}
}

\begin{document}
\maketitle

%%%%%%%%%%%%%%%%%%%%%%%%%%%%%%%%%%%%%%%%%%%%%%%%%%%%%%%%%%%%%%%%%%%%%%%%%%%%%%%%
\begin{abstract}
This paper presents an efficient learning-based method to solve the \textit{inverse kinematic} (IK) problem on soft robots with highly non-linear deformation. The major challenge of efficiently computing IK for such robots is due to the lack of analytical formulation for either forward or inverse kinematics. To address this challenge, we employ neural networks to learn both the mapping function of forward kinematics and also the Jacobian of this function. As a result, Jacobian-based iteration can be applied to solve the IK problem. A sim-to-real training transfer strategy is conducted to make this approach more practical. We first generate a large number of samples in a simulation environment for learning both the kinematic and the Jacobian networks of a soft robot design. Thereafter, a sim-to-real layer of differentiable neurons is employed to map the results of simulation to the physical hardware, where this sim-to-real layer can be learned from a very limited number of training samples generated on the hardware. 
%The effectiveness of our approach has been verified on pneumatic-driven soft robots for path following and interactive positioning.
\end{abstract}

% Keywords appear just beneath the abstract. Use only for final RAL version.  

\section{Introduction}
\label{sec:intro}
\IEEEPARstart{W}{ith} the use of flexible material, soft robots have the ability to make a large deformation and interact safely with the environment~\cite{Rus15_Nature}, which leads to a broad range of applications such as exoskeleton / wearable devices~\cite{Panagiotis15_RAS}, soft manipulators~\cite{Melingui15_TMECH, Drotman18_TMECH} and surgery assistance~\cite{Ranzani15_Bio}.    
However, as a hyper-redundant system with high nonlinearity in both material propriety and geometric deformation, it is difficult to formulate an effective kinematic model for solving the control task. The analytical \textit{forward kinematics} (FK) solution only exists for specific designs with a relatively simple shape (e.g.,~\cite{Gregory94_TRO,Jones06_TRO}). For a general soft robot with complicated structures / shapes, efficiently computing its IK solution remains a challenging problem. For soft robots with redundancy, fast and reliable IK solution is a very important means for improving the control precision and response frequency in practical tasks~\cite{Thuruthel18_SORO}.

\subsection{Related work} \label{subsec:relatedWork}
To efficiently model the behavior of soft robotic systems (i.e., computing FK), both analytical formulation and numerical simulation were conducted in previous research. Those analytical solutions, based on differential geometry~\cite{Gregory94_TRO,Jones06_TRO,Rolf12_IROS} and mechanics analysis~\cite{Renda14_TRO}, are difficult to be generalized for soft robots with a complex shape, where numerical simulation by the \textit{finite element method} (FEM) is usually employed~\cite{Moseley16_AEM, Xavier21_AIS}. Computational efficiency is a bottleneck of applying FEM in the IK computation, as the simulation needs to be repeatedly conducted to estimate the Jacobian~\cite{Orin_98IJRR}. To overcome this, a reduced model by voxel representation~\cite{Hiller14_SORO} or computing quasi-static equilibrium \cite{Duriez13_ICRA} are presented to accelerate. However, these methods can easily become non-realistic after applying large rotational deformation. The geometry-oriented simulation pipeline~\cite{Fang18_ICRA} can precisely compute the deformation of a variety of soft robots even in large rotation, which is later extended into a general IK solver~\cite{Fang20_TRO} by using the Jacobian-based iteration. A model reduction method is applied to further accelerate the numerical-based simulation~\cite{Goury18_TRO}. However, it is still difficult to directly include the simulator in the loop of iteration and achieve fast IK computing. 

\begin{table*}[ht]\footnotesize
\centering
\caption{Comparison of Learning-based Methods for Solving Inverse Kinematics on Soft Robots}\label{tab:IKLearningMethods}
\begin{tabular}{cc|c|c|c}  
\hline \hline
\specialrule{0em}{2pt}{1pt}
\multicolumn{2}{c|}{\multirow{2}{*}{\textit{}~}}  & \multicolumn{3}{c}{Learning-based Methods} \\  \specialrule{0em}{1pt}{2pt}
\cline{3-5}  \specialrule{0em}{1pt}{2pt}
\multicolumn{2}{c|}{Property} & \multicolumn{1}{c|}{Learning for IK Mapping~\cite{Giorelli13_IROS, Giorelli13_DSC, Rolf14_TNNLS, Giorelli15_TRO, Chen16_ICCAR,Grassmann18_IROS,reinhart17_sensors}}  & \multicolumn{1}{c|}{Jacobian by FK Network Gradient~\cite{Bern20_RoboSoft}} & \multicolumn{1}{c}{Learning Jacobian \& FK (Our work)}  \\  \specialrule{0em}{1pt}{2pt}
\hline \specialrule{0em}{1pt}{2pt}
%\multicolumn{2}{c|}{ \rev{Support for redundant system}{}}  & \rev{Require extra handling~\cite{reinhart17_sensors, Grassmann18_IROS}}{}  & \rev{Yes}{}  & \rev{Yes}{}  \\
\multicolumn{2}{c|}{Requirement on network type}   & General & Analytically differentiable network  & General \\
\multicolumn{2}{c|}{Smooth motion planning}     & Special extra effort needed~\cite{Grassmann18_IROS, reinhart17_sensors}  & Yes  & Yes \\
\multicolumn{2}{c|}{Target outside learning space}   & No  & Yes         & Yes \\ 
\multicolumn{2}{c|}{Converge speed} & No iteration  & Good     & Good  \\
\multicolumn{2}{c|}{Complexity of computation$^\dagger$} & $O(hb)$ & $O(hb^2)^\ddagger$ & $O(hb)$ \\ 
\specialrule{0em}{0pt}{2pt} \hline\hline
\end{tabular}
\begin{flushleft}
$^\dagger$We evaluate the complexity of network-based IK computing on networks with $O(h)$ hidden layers and $O(b)$ neurons per layer. \\
$^\ddagger$The high complexity in~\cite{Bern20_RoboSoft} is caused by applying the chain rule to a forward kinematic network to obtain its Jacobian, which results in nested functions. 
\end{flushleft}
\end{table*}

The data-driven methods used in soft robotics are often treated as regression problems of machine learning where kinematic models can be effectively learned from datasets~\cite{Thuruthel18_SORO}. To enable the inverse kinematic tasks on soft robots, an intuitive solution is to directly learn the mapping of IK which takes the motion as the input of a network and generates the corresponding parameters of actuation as output (ref.~\cite{Giorelli13_IROS, Giorelli13_DSC, Rolf14_TNNLS, Giorelli15_TRO,Chen16_ICCAR, reinhart17_sensors, Grassmann18_IROS}). However, this intuitive method does not perform well in a redundant system as the one-to-many mapping from task space to actuator space is generally difficult to learn. Although this issue can be partly solved by setting constraints in the actuator space~\cite{Grassmann18_IROS} or specifying the preference of configurations in the IK equation~\cite{reinhart17_sensors}, we solve the problem by using a different method to combine learning with Jacobian-based IK. Our method is efficient when planning a smooth motion (i.e., by minimizing variation in the actuator space) for soft robot systems with redundancy. To reach a similar goal, Thuruthel \textit{et al.}~\cite{Thuruthel16_ICAR, thuruthel16_SRDDC}
attempt to learn the differential IK model with local mapping. Another method is presented in~\cite{Li18_TMECH} to estimate the soft robot's Jacobian by Kalman filter approximation. Recently, Bern \textit{et al.}~\cite{Bern20_RoboSoft} presented a method to effectively evaluate the Jacobian by using the gradients of the FK network, which however limits the type of network used for FK learning and requires more time to compute the Jacobian for determining IK solutions. In our work, both the mapping functions of FK and the Jacobian are learned by neural networks as explicit functions. This makes our method far more efficient. A comparison of three types of learning-based methods is given in Table \ref{tab:IKLearningMethods}.

On the other hand, learning a kinematic model for soft robots usually needs a large number of samples, which can be very time-consuming when generating the data in a physical environment either by the motion capture system~\cite{Drotman18_TMECH} or embedded sensors~\cite{Thuruthel19_scirobot, Scharff19_TMECH}.  Moreover, to explore the boundary of the work space, a large extension in material under large actuation needs to be applied \cite{Sun13_IROS}. Soft materials on a robot can become fragile and might generate plastic deformation after repeating such deformation may times~\cite{Drotman18_TMECH}. Consequently, the learned model becomes inaccurate. Furthermore, errors generated during the fabrication of a specimen can make the network learned on this specimen difficult to be used on other specimens with the same design.
To reduce the cost of generating training data, Daniel \textit{et al.} reported a data-efficient method by exploiting structural properties of the kinematic mapping~\cite{Kubus18_IROS}. Another solution is to generate accurate dataset in the simulation environment and then convert the model learned from simulation into physical reality by transfer learning (ref.~\cite{Marquez18_TNNLS, Kriegman20_RoboSoft, Park20_RAL, Donat20_TMRB}). The hybrid model contains the analytical formulation and the network-based correction is conducted in~\cite{Malekzadeh14_RoBio, reinhart17_sensors} where more precise control of the soft robot is achieved. Similarly, FEM is used in~\cite{Mats21_RoboSoft} to generate a simulation dataset for training a hybrid kinematic model by transfer learning. A more efficient numerical simulator~\cite{Fang20_TRO} is adopted in this work to generate the training dataset. When working together with the sim-to-real network, IK with high accuracy can be achieved. The comparison of IK with sim-to-real learning by using the reduced analytical model vs. our simulation based model can be found in Sec.~\ref{subsec:discussion}.

\subsection{Our Method}

\begin{figure} [t]
\centering
\includegraphics[width=\linewidth]{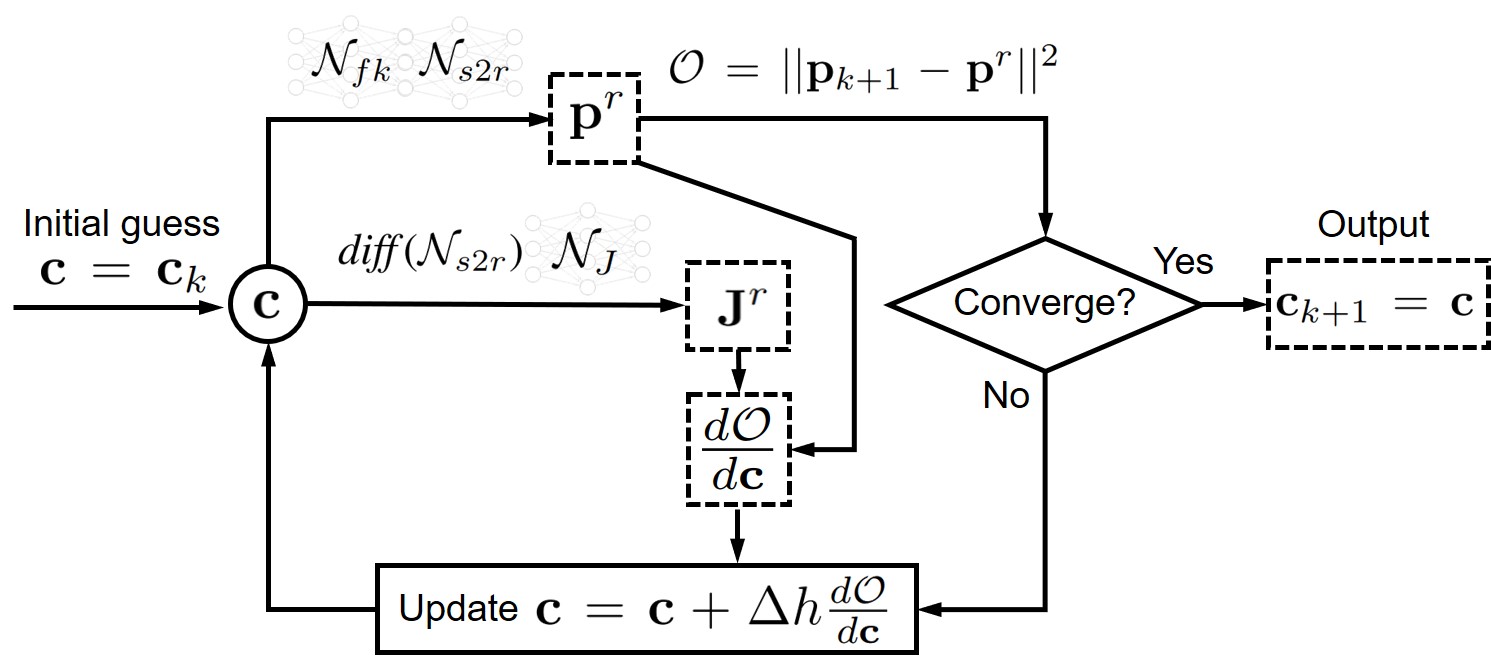}
\caption{Pipeline of Jacobian-based method to determine the actuation parameters $\mathbf{c}_{k+1}$ for the way-point $\mathbf{p}_{k+1} \in \mathcal{L}$ that solves the IK problem of soft robots by minimizing $\mathcal{O}(\cdot)$ in (\ref{eq:objectives}). Both the position $\mathbf{p}^r$ in task space and the Jacobian $\mathbf{J}^r$ are effectively estimated by the networks $\mathcal{N}_{fk}$, $\mathcal{N}_{J}$ and $\mathcal{N}_{s2r}$ obtained from offline training.
When $\mathcal{O} < \epsilon^2$,
%When $\| \mathbf{p}_{k+1} - \mathbf{p}^r\| < \epsilon$, 
we regard the convergence as having been achieved (e.g., 0.1\% of the workspace width is employed as $\epsilon$).}
\label{fig:Algorithmframework}
\end{figure}

Three networks - 1) forward kinematics $\mathcal{N}_{fk}$, 2) Jacobian $\mathcal{N}_{J}$, and 3) sim-to-real mapping $\mathcal{N}_{s2r}$ are trained to support the effective computing of IK for soft robots in both virtual and physical spaces at a fast speed. With an objective function defined in quadratic form and the network-based efficient estimation of FK and Jacobian, Jacobian-based iteration is used to compute the IK solution. The pipeline of our method is presented in Fig.~\ref{fig:Algorithmframework} with detail discussed in Sec.~\ref{sec:Kinematics}. 

The technical contributions of our work are:
\begin{itemize}
\item A direct pipeline for learning both the FK and Jacobian from accurate numerical simulation results, to support effective IK computing for soft robots. This method can compute IK solutions in a fast speed and has the capability to plan a smooth motion for soft robotic systems with redundancy.

\item A two-step learning strategy by using the sim-to-real transfer learning to eliminate the gap between the prediction based on simulation and the physical behavior, which can greatly reduce the required amount of empirical data when compared to directly learning a predictor from the physical experiment.

\end{itemize}
The behavior of our method has been verified on two hardware setups of soft robots giving 2D and 3D motions. The effectiveness of our method is quantitatively evaluated and compared with other approaches in the IK tasks of soft robots. Experimental tests are also conducted to demonstrate the performance of our method on soft robots with the same design but fabricated with different materials. 

\section{Jacobian-Based Kinematics and Learning}
\label{sec:Kinematics}
In this paper, we focus on solving the IK problem for soft robots -- specifically, to determine the parameters of actuation that can drive a soft robot to reach a target position / shape. As the analytical IK solution cannot be obtained, we adopt a Jacobian-based numerical method where a target-oriented objective function $\mathcal{O}(\cdot)$ is minimized to determine the parameters of actuation. In this section, we first introduce the Jacobian-based IK computation for the path following task. After that, we demonstrate how it can be solved practically by applying the training in a virtual environment and then the sim-to-real transformation. 

\subsection{Jacobian-based IK solution}\label{subsec:JacobianBasedIK}
The path following problem of a soft robot is described as driving a marker on its end-effector to move along a path $\mathcal{L}$ presented by a set of target waypoints $\{ \mathbf{p}_1, \mathbf{p}_2, \cdots, \mathbf{p}_{i}, \mathbf{p}_{i+1}, \cdots \}$ in the task space. For each waypoint $\mathbf{p}_{i}$ to be reached by the marker, numerical computation of inverse kinematics attempts to minimize the distance between $\mathbf{p}_{i}$ and the marker's position. This is formulated as an optimization problem
\begin{equation}
    \mathbf{c}_i = \argmin_{\mathbf{c}} \mathcal{O}(\mathbf{p}_i, \mathbf{c}) = \argmin_{\mathbf{c}} \| \mathbf{p}_i - \mathbf{p}(\mathbf{c}) \|^2
    \label{eq:objectives}
\end{equation}
where $\mathbf{p}(\cdot) \in \mathbb{R}^n$ denotes the forward kinematic function to compute the position of the marker. The input of $\mathbf{p}(\cdot)$ is a vector of actuation parameters, $\mathbf{c} = (c_1,c_2, \cdots, c_m) \in \mathbb{R}^m$. Here $n$ and $m$ are dimensions of the task space and the actuator space respectively. 

To find the solution of (\ref{eq:objectives}), the gradient of $\mathcal{O}(\cdot)$ as
\begin{equation}\label{eq:gradients}
    \frac{d\mathcal{O}}{d\mathbf{c}} = -2( \mathbf{p}_i - \mathbf{p}(\mathbf{c})) \mathbf{J}(\mathbf{c}) 
\end{equation}
will be employed to update the value of $\mathbf{c}$ with $ \mathbf{J}(\mathbf{c}) = d\mathbf{p}/d\mathbf{c} \in \mathbb{R}^{n \times m}$ being the Jacobian matrix that describes the moving trend of a soft robot's body at certain actuation parameters. The value of $\mathbf{c}$ is updated by $\mathbf{c}=\mathbf{c}+\Delta h \frac{d\mathcal{O}}{d\mathbf{c}}$. $\Delta h$ is a step size to minimize the value of $\mathcal{O}(\cdot)$ along the gradient direction, which can be determined by soft linear-search \cite{Fang20_TRO}. 
%(detail can be found in \cite{Fang20_TRO} and the supplementary document). 
Fig.~\ref{fig:Algorithmframework} gives the illustration of this algorithm.

When a physics-based simulation is employed to evaluate the forward kinematic function $\mathbf{p}(\cdot)$, the Jacobian matrix $\mathbf{J}$ can be obtained by numerical difference \cite{Fang20_TRO, Mats21_RoboSoft}. The $k$-th column of $\mathbf{J}$ is computed as
\begin{equation}\label{eq:numericalDiff}
    \mathbf{J}_k = \frac{\partial \mathbf{p}(\mathbf{c})}{\partial c_k}  \approx  \frac{\mathbf{p}(..., c_k+\Delta c, ...) - \mathbf{p}(..., c_k-\Delta c, ...)}{2 \Delta c},
\end{equation}
where $\Delta c$ is a small constant determined according to experiments and assigned as $1/10N$ of the actuation range. Here $N$ is the number of samples for each actuation parameter presented in Sec.~\ref{subsec:dataGenerationSim}. Notice that it can be time-consuming to evaluate the values of $\mathbf{p}(\cdot)$ and $\mathbf{J}(\cdot)$ by physics-based simulation in IK computing. We therefore introduce a learning-based method to learn both the FK and the Jacobian model in the offline stage, which can support a fast IK computing during online usage. In the meantime, the difference between the simulation and the physical behavior is fixed by the sim-to-real transfer learning. 
\subsection{Learning-based model for IK computing}\label{subsec:networkForIK}

\begin{figure}[t]
\centering
\includegraphics[width=0.9\linewidth]{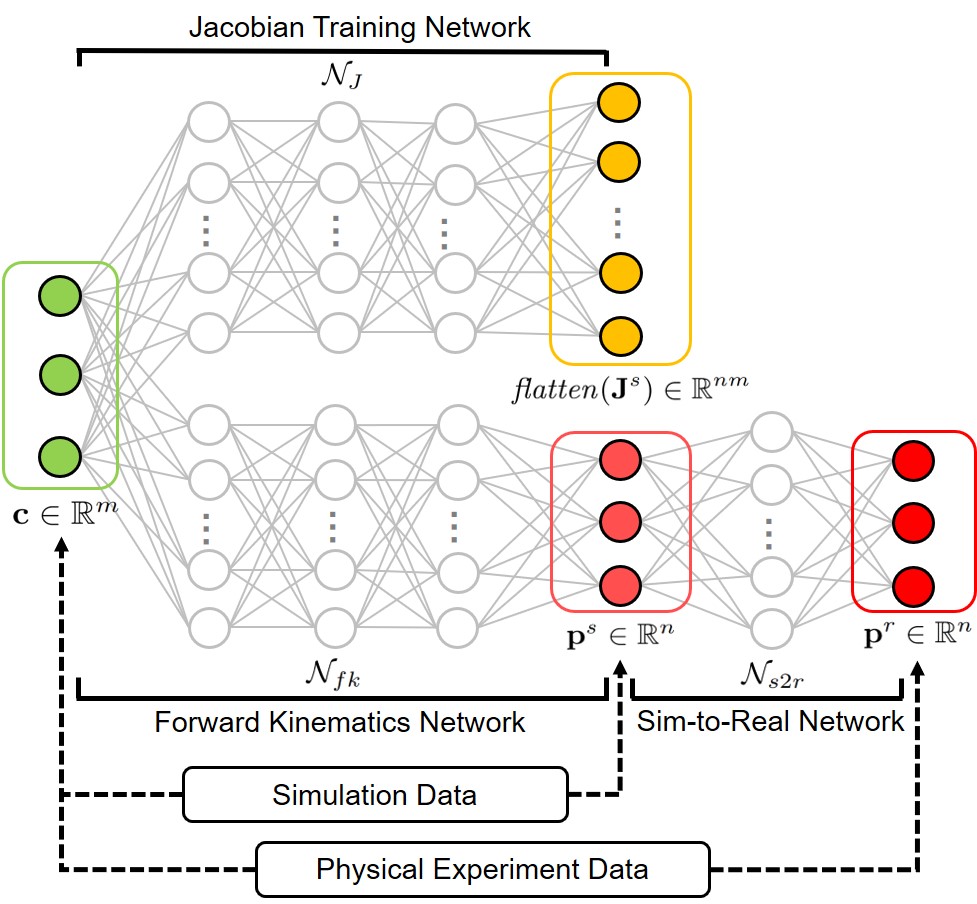}
\caption{The network structure used in our approach to train the kinematic model and the sim-to-real transfer.}
\label{fig:network}
\end{figure}

\begin{figure*} [t]
\centering
\includegraphics[width=\linewidth]{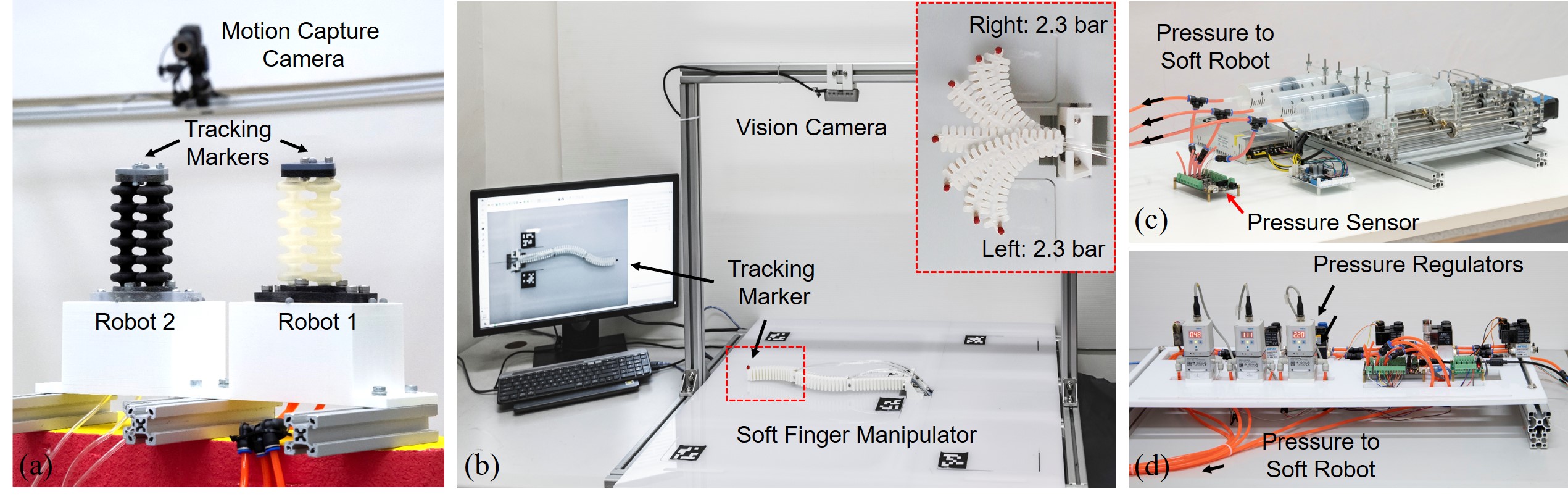}
\caption{Two hardware setups employed in our experiments to collect data and verify the performance of our method -- (a) a soft actuator with multiple chambers that are actuated by an array of syringes (see (c)) and (b) three connected soft fingers that can be actuated individually by proportional pressure regulators (see (d)).}\label{fig:setup}
\end{figure*}

We learn both the forward kinematic model and its Jacobian from simulations -- denoted by $\mathbf{p}^s(\cdot)$ and $\mathbf{J}^s(\cdot)$, which are transferred to physical hardware by learning a sim-to-real mapping function $\mathbf{r}(\cdot)$. Denoting the location of a traced marker on physical hardware as $\mathbf{p}^r$, the function of sim-to-real mapping is required to have $\mathbf{r}(\mathbf{p}^s) \approx \mathbf{p}^r$. Neural networks are employed to learn these functions (see the architecture of neural networks shown in Fig.~\ref{fig:network}).

In the simulation environment, $\mathbf{p}^s(\cdot)$ and $\mathbf{J}^s(\cdot)$ are trained on two networks $\mathcal{N}_{fk}$ and $\mathcal{N}_{J}$ by spanning the work space of actuators with a large number of samples. Note that the output layer for $\mathcal{N}_J$ is a column vector as the flattened Jacobian matrix $\mathbf{J}^s$. After obtaining the network $\mathcal{N}_{fk}$, the sim-to-real mapping function $\mathbf{r}(\cdot)$ is trained on a differentiable network $\mathcal{N}_{s2r}$ by using a few samples obtained from a physical experiment conducted on the hardware setup. 

With the help of these trained networks, we can estimate the Jacobian on the hardware setup as
\begin{equation}
  \mathbf{J}^r(\mathbf{c}) = \frac{d\mathbf{p}^r}{d\mathbf{p}^s}\frac{d\mathbf{p}^s}{d\mathbf{c}}  \approx  
  %\frac{d\mathbf{r}}{d\mathbf{p}^s}\frac{d\mathbf{p}^s}{d\mathbf{c}} = 
  \mathrm{diff} (\mathcal{N}_{s2r}) \mathbf{J}^s(\mathbf{c})
\end{equation}
Considering the difficulty of data acquisition on hardware specimens, the \textit{feed-forward neuronal network} (FNN) with a single layer of fully connected neurons is adopted in our implementation for $\mathcal{N}_{s2r}$. The differentiation $\mathrm{diff} (\mathcal{N}_{s2r})$ as a $n \times n$ matrix can be computed analytically by differentiating the network's activation functions. As most of the complexity in kinematics can be effectively captured by $\mathcal{N}_{fk}$ and $\mathcal{N}_{J}$, a lightweight network $\mathcal{N}_{s2r}$ trained by a small dataset obtained from the physical experiment can already show very good performance on eliminating the inconsistency in material properties and fabrication.

Through this learning-based model, the gradient of the IK objective function in the physical environment can then be computed by
\begin{align}
    \frac{d\mathcal{O}}{d\mathbf{c}} &= -2( \mathbf{p}_i - \mathbf{p}^r(\mathbf{c})) \mathbf{J}^r(\mathbf{c}) \label{eq:realJacobian} \\
    &\approx -2( \mathbf{p}_i - \mathbf{r}(\mathbf{p}^s(\mathbf{c}))) \mathrm{diff} (\mathcal{N}_{s2r}) \mathbf{J}^s(\mathbf{c})   \label{eq:leanredJacobian}
\end{align}
Note that the real positions of markers, $\mathbf{p}^r(\mathbf{c})$ in (\ref{eq:realJacobian}), can also be obtained from a hardware setup (e.g., by a motion-capture system). However, using positions predicted by $\mathcal{N}_{fk}$ and $\mathcal{N}_{s2r}$ networks can avoid physically actuating the hardware inside the loop of numerical iteration. After training the networks $\mathcal{N}_{fk}$, $\mathcal{N}_{J}$ and $\mathcal{N}_{s2r}$, an iteration-based algorithm (as shown in Fig.~\ref{fig:Algorithmframework}) is used to effectively solve (\ref{eq:objectives}). The actuation parameters $\mathbf{c_{i-1}}$ for realizing $\mathbf{p}_{i-1}$ are employed as the initial guesses when computing $\mathbf{c_i}$ for $\mathbf{p}_{i}$. As a result, the iteration converges rapidly and the continuity of motion in configuration space can be preserved. 

\section{Data Generation and Training}\label{sec:data_training}
We first present two hardware setups that are used in our research to verify the performance of the learning-based method presented above. After introducing the steps for generating datasets, the training details are provided.

\subsection{Soft robotic hardware}\label{subsec:twoSoftRobSystems}
Two hardware setups are constructed to investigate the performance of our IK solver. Both setups are equipped with cameras to capture the real positions of markers for the purpose of training and verification. 

\subsubsection{Actuator with 3D motion} The first setup is a 3D printed soft actuator with three chambers that can be actuated individually~\cite{Drotman18_TMECH}. Its soft body can extend and bend in a 3D task space. To verify the behavior of our sim-to-real method, two specimens are fabricated by the same Object 350 Connex 3D printer but using slightly different materials -- the Agilus Black and Agilus transparent materials. Both have the softness 70A according to their factory specification. These two models are shown as Robot 1 and Robot 2 in Fig.~\ref{fig:setup}(a). The soft robot is actuated by an array of syringes that has close-loop control with the help of pressure sensors as shown in Fig.~\ref{fig:setup}(c). For this setup, we have the same dimension for the work space ($m=3$) and the actuator space ($n=3$).

\subsubsection{Planar finger manipulator} The second setup is a soft manipulator that can move in the $xy$-plane (see Fig.~\ref{fig:setup}(b)). The manipulator contains three soft finger sections that are rigidly connected. We use Festo Pressure Regular VPPE-3-1/8-6-010 to provide the pressure for each section (see Fig.~\ref{fig:setup}(d)). Every finger section contains dual chambers that can bend symmetrically for both sides up to $120^{\circ}$. %(ref.~\cite{Yap16_SORO}). 
To maximize the deformation of each finger section, we only actuate one side for a segment each time with the pressed air in the range of $[0 ,3]$ bar. When considering both sides of a segment, this results in a range of $[-3,3]$ as actuation -- i.e., `+' for actuating the chamber at one side and `-' for the other side. This is a redundant system with the dimension $n=2$ for the work space and $m=3$ in the actuator space.

\subsection{Data generation on simulator}\label{subsec:dataGenerationSim}
In our work, forward kinematics of soft robots in a virtual environment is computed by a geometry-oriented simulator \cite{Fang18_ICRA, Fang20_TRO}.
%
\iffalse
\rev{~which outperforms in its efficiency and capability to handle large rotational deformation.}{, where soft robot body and chamber region are represented by tetrahedral meshes $\mathcal{M}_s$ and $\mathcal{M}_c$, respectively. Once the actuation (i.e., volume expanding ratio $\alpha$ in chamber region) was applied, the system trends to computes the deformed shape $\mathcal{M}_s^d$ of soft robot by minimizing the geometry-based elastic energy $E_{elastic}$.}
\begin{align}
\argmin_{\mathcal{M}_s^d} \quad & E_{elastic} = \sum_{e \in \mathcal{M}_s} \mathit{Vol}(e) ||\mathbf{N}\textbf{V}_e^d -\mathbf{R} (\mathbf{N}\textbf{V}_e)||^2  \label{eq:eleEnengy} \\
s.t. \quad & \alpha \mathit{Vol}(\mathcal{M}_c) - \mathit{Vol}(\mathcal{M}_c^d) = 0. \label{eq:actuConst}
\end{align}
\rev{}{Here $\mathbf{V}_{e} \in \mathbb{R}^{3 \times 4}$ describe the shape of a tetrahedron element $e$ by its four vertex position. $\mathbf{V}_{e}$ and $\mathbf{V}_{e}^d$ give the initial and deformed shape respectively. The function $\mathit{Vol}(\mathcal{M}_c)$ computes the volume of the chamber region. $\mathbf{N}$ and $\mathbf{R}$ are matrices used to align elements into same coordinate. This constrained optimization problem is solved effectively by a local-global iterative solver~\cite{Fang20_TRO}, which shows good convergence on soft robots with large rotational deformation (e.g., soft finger in Fig.~\ref{fig:setup}(b)).}
\fi
%
When employ this simulator to generate datasets for training the forward kinematic network $\mathcal{N}_{fk}$ and the Jacobian network $\mathcal{N}_J$, the computation time for generating single sample point is $4.3$ sec. (the three-chamber robot) and $1.2$ sec. (the finger manipulator) respectively. 
It's worth mentioning that as a general training pipeline, the dataset used to train $\mathcal{N}_{fk}$ and $\mathcal{N}_{J}$ can be generated by different kinematic models -- e.g., those analytically computed by piecewise constant curvature (PCC)~\cite{Melingui15_TMECH} or numerically by FEM software such as Abaqus. Here we choose the geometry-based simulator as it can further reduce the cost of data generation than FEM. Moreover, it needs to capture less physical data than the analytical model for training the sim-to-real network (discussed Sec.~\ref{subsec:discussion}).

\begin{figure} [t]
\centering
\includegraphics[width=\linewidth]{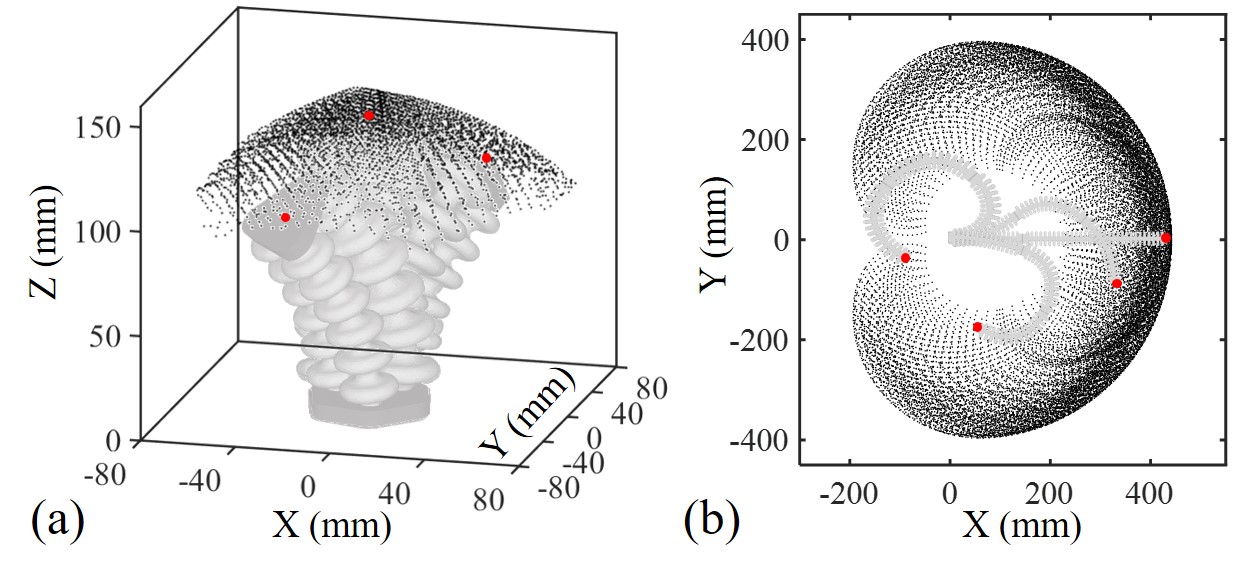}
\caption{The results of the simulation are employed as training samples (present in black dots) to learn the forward kinematic network $\mathcal{N}_{fk}$ and the Jacobian network $\mathcal{N}_J$, where these samples also span the workspace $\mathcal{P}^w$ of a robot. Red dots represent some example points as targets for motion in the workspace.
}
\label{fig:FKsimualtion_workspace}
\end{figure}

We now present the sampling method in actuator space for generating training data points. We uniformly divide the pressure range of each actuator into $N$ segments to make sure that the distance between sample points is less than $1\%$ of the workspace width. Sampling results of the two hardware setups are shown in Fig.~\ref{fig:FKsimualtion_workspace}, which also presents the workspaces $\mathcal{P}^w$ of these two soft robots. In our experiment, $N=16$ and $N=29$ are used for these two setups respectively. This results in $16^3 = 4096$ samples for the three-chamber actuator (Fig.~\ref{fig:FKsimualtion_workspace}(a)) and $29^3 = 24389$ samples for the finger manipulator (Fig.~\ref{fig:FKsimualtion_workspace}(b)). Notice that the difference in 
choosing the $N$ value is due to the redundancy of the finger setup, which also has a larger range in actuation. Based on our tests, datasets selected in these sizes can already well train $\mathcal{N}_{fk}$ and $\mathcal{N}_{J}$ to capture the kinematic behavior of soft robots (see the results in the following section).

\subsection{Data generation on hardware}\label{subsec:dataGenerationHardware}
For the purpose of training sim-to-real network $\mathcal{N}_{s2r}$, datasets are generated on two hardware setups.
We uniformly span the actuator space to generate physical data which are classified into the training ($70\%$) and the test ($30\%$) dataset.
Since the efficiency of the training pipeline depends on the number of samples generated on hardware setups, we test and determine the appropriate sample number used to train $\mathcal{N}_{s2r}$ -- details are presented in Sec.~\ref{subsec:trainingDetails}.

\subsubsection{Actuator with 3D motion} To trace the 3D motion of this soft actuator, we place a marker at the center of its top plane and several markers on its static base. The motion capture system that contains 8 Vicon Bonita 10 cameras and 10 Vicon Vantage 5 cameras is used to capture the movements at the rate of 30 Hz. Because of the viscoelasticity of soft materials used to fabricate this robot, it takes a relatively long time for the position of a marker to become stable (i.e., less than 0.05mm change between neighboring image frames). This makes the process of data collection more time-consuming than a robotic system with rigid bodies. As a result, the average time for collecting one sample in the physical environment is $4.0$ sec.

\subsubsection{Planar finger manipulator} As only planar coordinates are needed when tracking the positions of a marker, we use a RealSense D435 camera mounted at the top of the setup. We place a red marker on the tip of the manipulator and adopt the OpenCV library as software to track the marker's position in the plane. QR code is employed to build the mapping between the coordinates in image space and the coordinates in the real world. The speed of data acquisition for this system is 10 Hz. For this hardware setup, the average time for collecting one sample point is $3.5$ sec.
%\guoxin{we may need to add the reason why different actuation setup is used here}

\subsection{Details of training}\label{subsec:trainingDetails}

\begin{figure}[t]
\centering
\includegraphics[width=\linewidth]{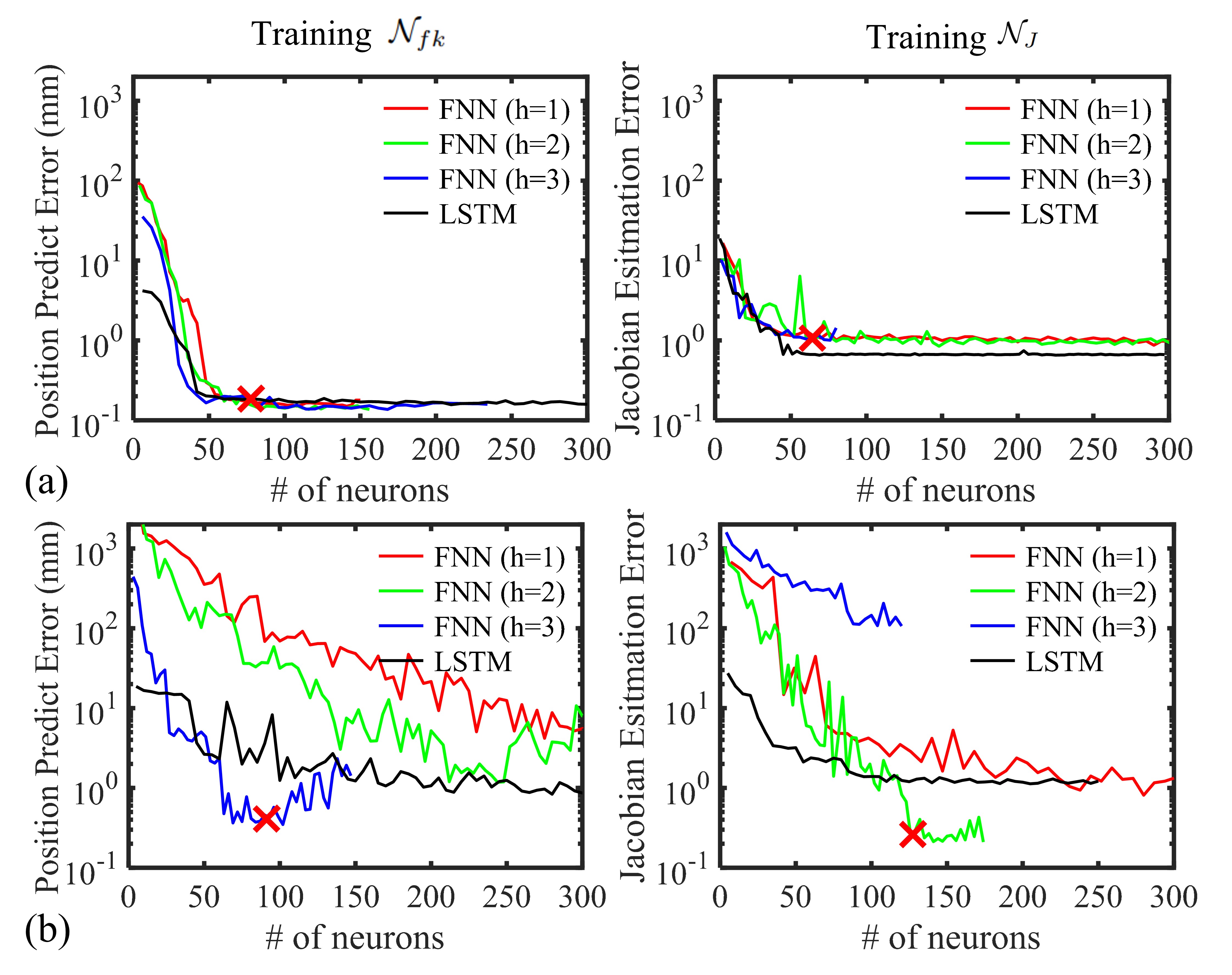}
\caption{Comparison of learning results by using a different number of layers as $h$ and a different number of neurons $b$ per layer (i.e., the total number of neurons in a network is $hb$). Tests are conducted on (a) the robotic setup without redundancy (the three-chamber actuator with $m=n=3$) vs. (b) the setup with redundancy (the planar finger manipulator having $m=3$ and $n=2$). Red crosses indicate the network parameters used in the physical experiment.
}\label{fig:training_detail}
\end{figure}

In this work, the type and structure of training models used in experiment are carefully selected based on their performance. For training $\mathcal{N}_{fk}$ and $\mathcal{N}_J$, FNN and \textit{Long Short-term Memory} (LSTM) are tested as they can both adequately capture the nonlinear behavior in a training dataset, including the many-to-one forward kinematic mapping for redundant systems. For the sim-to-real transfer network $\mathcal{N}_{s2r}$, it needs to be differentiable with analytic gradients. Meanwhile, it should be lightweight as only a limited number of samples can be obtained from physical experiments. 
For these reasons, single-layer FNN is selected for $\mathcal{N}_{s2r}$.
All networks are trained by using the Deep Learning Toolbox of MATLAB running on an NVIDIA GeForce RTX 2070 graphics card. 

\subsubsection{Training for FK and Jacobian}
We first study the effectiveness of training $\mathcal{N}_{fk}$ and $\mathcal{N}_{J}$ by using a different number of layers and different numbers of neurons.
Each data set is divided into training and test subsets in the ratio of $70\%:30\%$. For all networks, the activation function is set as Tan-Sigmoid
\begin{equation}
    f(x) = \frac{2}{1+e^{-2x}}-1
    %f(x) = -1+{2}/{(1+e^{-2x})}
\end{equation}
as it can well fit the non-linearity in kinematic mapping. Moreover, it's differentiable and can provide a faster training speed. We set the batch size as 200, the maximum number of epochs as $13500$ and the learning rate as $0.04$. The Levenberg-Marquardt backpropagation is employed for training.

The estimation errors for both soft robot setups are evaluated on the test datasets as shown in Fig.~\ref{fig:training_detail}, where we can find that FNN and LSTM can both converge to a good result after carefully tuning the network parameters. It is hard to find significant improvement in the accuracy by using network with feedback connections (i.e., LSTM). This is because the training data only contain quasi-static information of the system, where the time-related network structure cannot show its advantage. On the other hand, it is found that the structure of the network for learning the Jacobian $\mathcal{N}_{J}$ on a redundant system (i.e., the planar finger manipulator) needs to be selected more carefully. 
FNN is selected as the final network structure, and the best performance for training $\mathcal{N}_{J}$ is observed on this hardware setup when FNN with $h=2$ hidden layers and $b = 64$ neurons per layer is employed to learn $\mathcal{N}_{J}$. Differently, FNN with $h=3$ layers and $b = 30$ neurons per layer gives the best results in training $\mathcal{N}_{fk}$. The error of position prediction by using $\mathcal{N}_{fk}$ is less than $0.5 \ \mathrm{mm}$ (i.e., $0.58\%$ of the work space's width). For the three-chamber actuator, the numbers of layers and neurons have less influence on the training result. For this setup, we select $h=2$ and $b=35$ for both networks, which results in a FK prediction with error less than $0.17 \ \mathrm{mm}$ on the test dataset (i.e., $0.34\%$ of the workspace's width). With such accurate predictions generated by $\mathcal{N}_{fk}$ and $\mathcal{N}_{J}$, we can obtain IK solutions efficiently and accurately (see the behavior study given in Section~\ref{sec:experimentResult}).

\subsubsection{Training for sim-to-real transfer} 
\begin{figure}[t] 
\centering
\includegraphics[width=\linewidth]{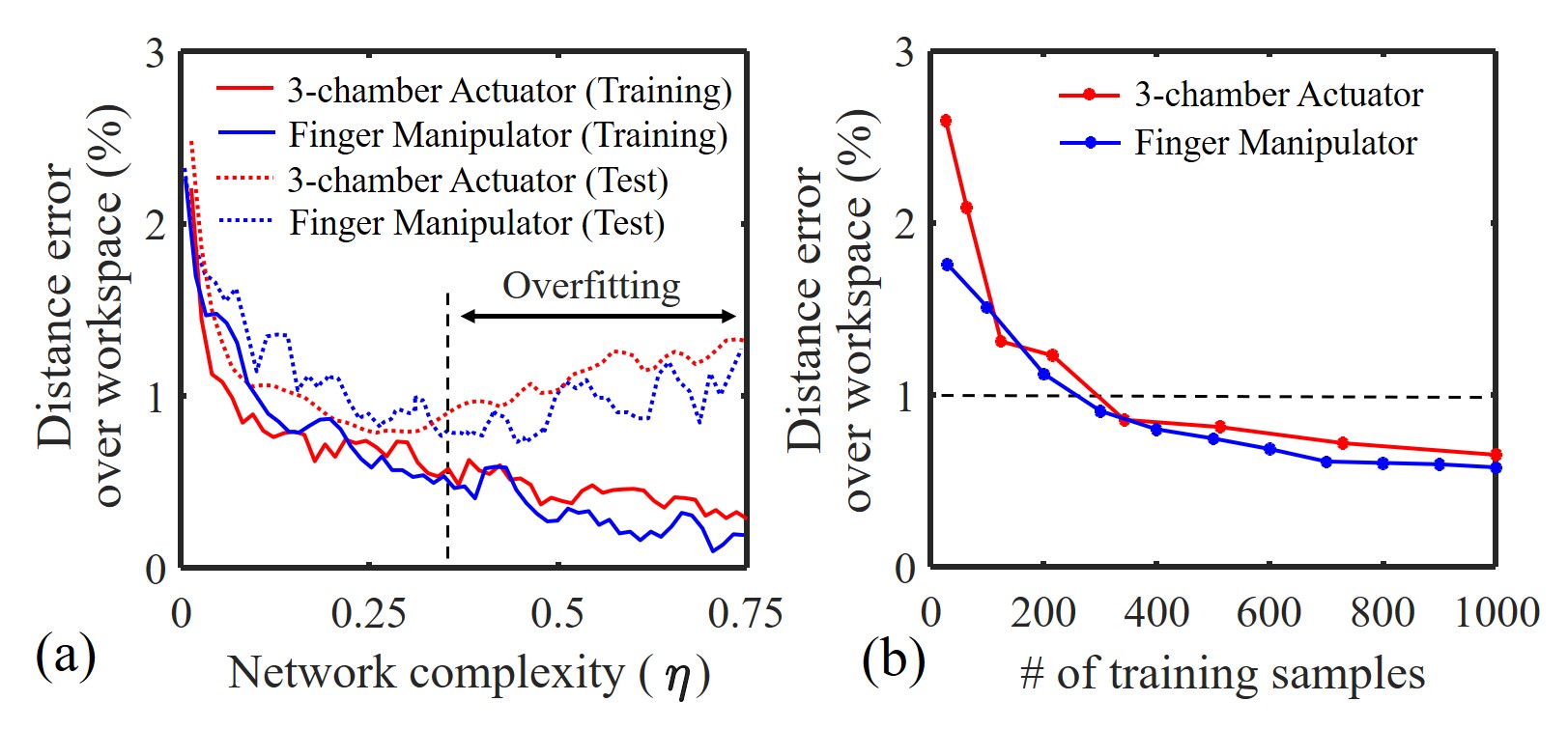}
\caption{Experimental study for the performance influence in the sim-to-real network $\mathcal{N}_{s2r}$ by using (a) different numbers of neurons and (b) datasets in different sizes. With the properly selected complexity of network structure, the over-fitting problem can be avoided. The distance predict error can be controlled within $1\%$ of the workspace width for both setups when a limited number of training samples are used (see (b)).}\label{fig:sim2realTraining}
\end{figure}

When training for $\mathcal{N}_{s2r}$ an important parameter here is the number of neurons, which is selected as $\eta$ times the number of samples to avoid over-fitting on the training dataset. Fig.~\ref{fig:sim2realTraining}(a) presents the behavior on both the training dataset (denoted by the solid curves) and the test dataset (denoted by the dash curves) when using different values of $\eta$. Based on the analysis, $\eta = 1/4$ is selected for our experiment to avoid over-fitting.

As the time used to collect physical data points should be controlled, we also study the behavior of $\mathcal{N}_{s2r}$ with different numbers of training samples. For this purpose, the prediction errors as the ratios of the distance errors over the workspace widths are given in Fig.~\ref{fig:sim2realTraining}(b) to study the effectiveness of using different numbers of samples. In these tests, the number of neurons is always assigned as $\eta = 1/4$ of the training samples. For both setups, we find that the network $\mathcal{N}_{s2r}$ can be well trained when using a limited number of training samples. In our implementation, $1 \%$ is selected as the threshold of accurate prediction and this threshold is used to determine the number of samples for training $\mathcal{N}_{s2r}$. As a result, $343$ samples are used for the three-chamber actuator and $620$ samples are conducted for the finger manipulator. The datasets for training $\mathcal{N}_{s2r}$ on two hardware setups can both be collected within $30~\mathrm{min}$.

\section{Experiment Results}\label{sec:experimentResult}
In this section, we present all the experiment results of IK computing for soft robots by using our learning based Jacobian iteration. The results are generated in both the virtual and the physical environments. Computation of the learned neural networks in prediction is implemented in C++ and integrated into our control platform to gain the best computational efficiency. All the IK computations were efficiently run on a laptop PC with Intel i7-9750H 2.60GHz CPU and 16GB memory. Note that the prediction made by networks and the IK algorithm is entirely run on a CPU.

\subsection{Path following by actuator with 3D motion} \label{subsec:trajFollow}

\begin{figure}[t]
\centering
\includegraphics[width=\linewidth]{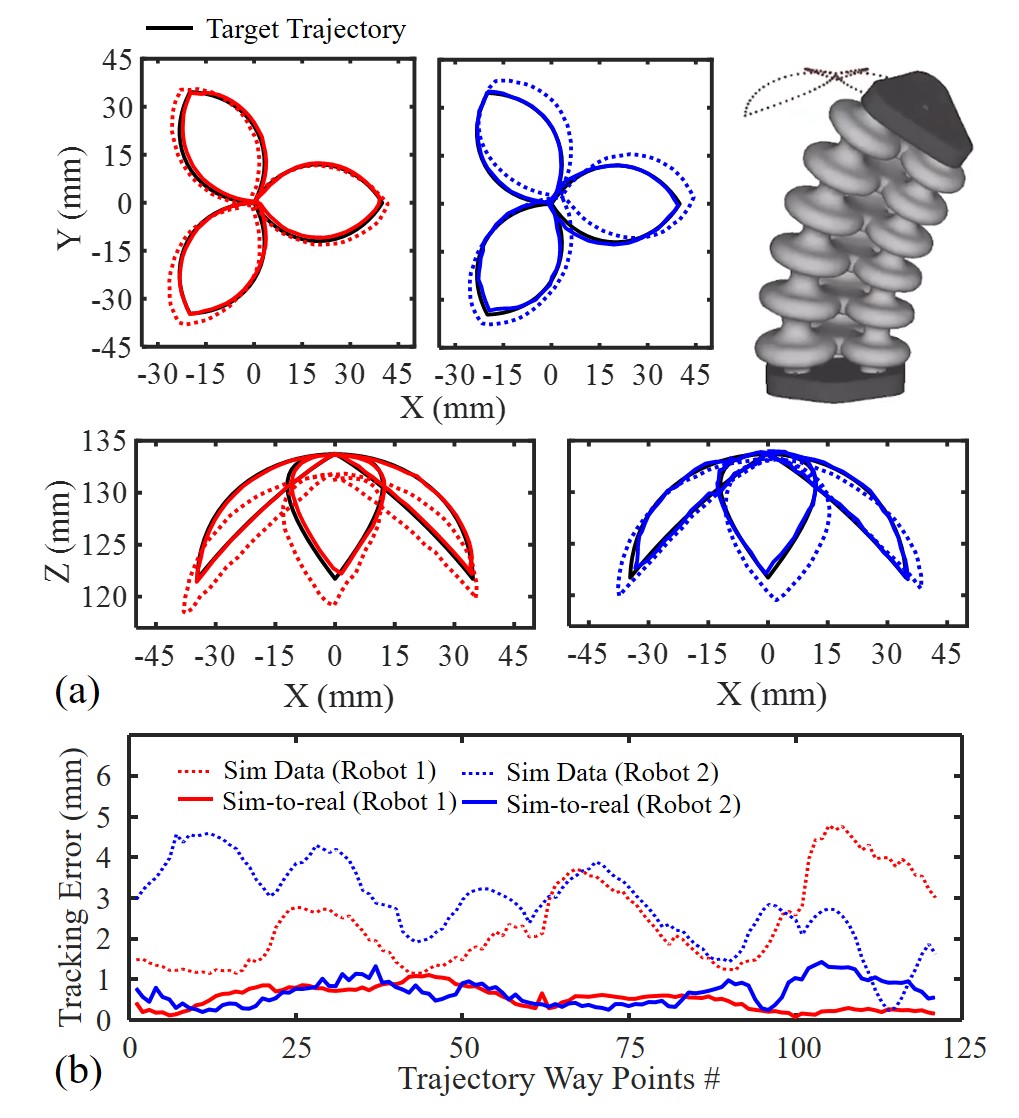}
\caption{Results of path following task on two soft robots with the same design but fabricated with different materials (shown as Robot 1 and Robot 2 in Fig.~\ref{fig:setup}(a)). (a) Trajectory of the soft robots by applying IK solutions with (solid line) and without (dash line) the sim-to-real network. (b) Visualized tracking errors on trajectory waypoints.
}\label{fig:three_chamber_result}
\end{figure}

\begin{figure}[t]
\centering
\includegraphics[width=0.9\linewidth]{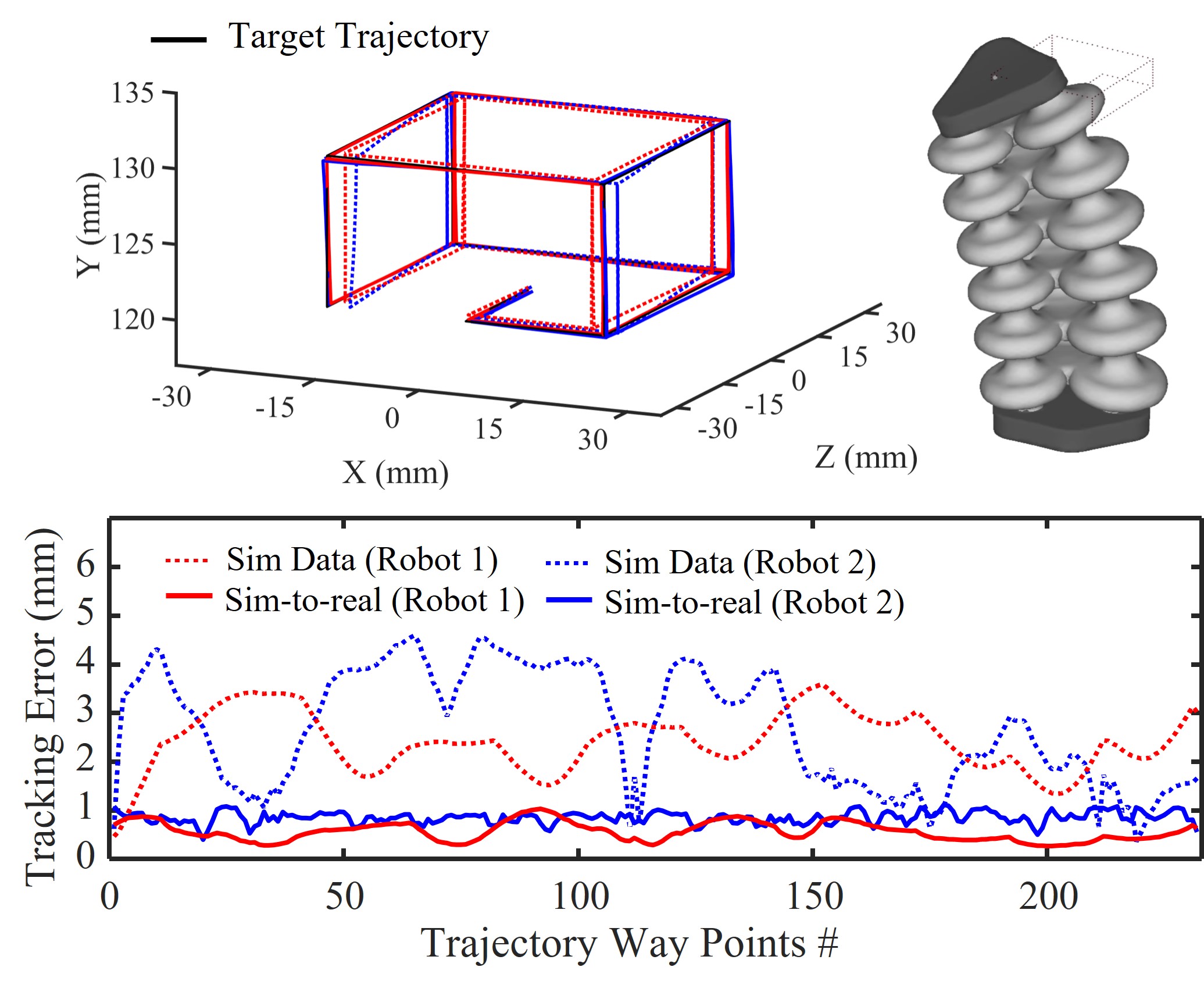}
\caption{Results of path following for the 'box' shape trajectory demonstrate the vertical motion of three-chamber soft robot. It can also find that the maximal tracking errors are reduced from $4.5\mathrm{mm}$ to $1.0\mathrm{mm}$ for both robots after applying the sim-to-real network.
}\label{fig:three_chamber_vertical_traj}
\end{figure}

%\subsubsection{Actuator with 3D motion}
We first test the behavior of the learning-based IK computing method in the task of path following on the soft actuator with three chambers. 
Given 3D trajectories of the `flower' shape (Fig.~\ref{fig:three_chamber_result}) and the `box' shape (Fig.~\ref{fig:three_chamber_vertical_traj}), 120 and 240 waypoints are sampled on the paths in uniform distances respectively. The proposed IK solver is then used to compute the actuation parameters for each waypoint. This leads to the actuation sequence that can drives soft robot to move along the trajectories.
When running in the simulation environment, the trained networks can generate actuation that results in very accurate trajectories with the average tracking error as $0.13\mathrm{mm}$. 
In the physical environment, we learned the sim-to-real networks separately on two soft robots as shown in Fig.~\ref{fig:setup}(a). If we directly apply the actuation parameters obtained from IK computing in the simulation environment, the error of path following is high (i.e., up to $5 \  \mathrm{mm}$). At the same time, the variation caused by fabrication and material can be clearly observed from the difference between Robot 1 and 2, as shown in Fig.~\ref{fig:three_chamber_result}. By incorporating the sim-to-real transfer in our method, we can successfully reduce the error in the physical environment to less than $1.2\mathrm{mm}$ for both robots (see Fig.~\ref{fig:three_chamber_result}(b)), which is $1.71\%$ of the workspace width. For the tests given in Fig.~\ref{fig:three_chamber_vertical_traj}, the maximal errors are reduced from $4.5\mathrm{mm}$ to $1.0\mathrm{mm}$. It's interesting to see that tests with circular and line trajectories shown similar tracking error. This is because the IK mapping from task space to actuator space is nonlinear.

\begin{table}[t]
\centering
\caption{Computing Speed and IK Accuracy by Different Methods}
\begin{tabular}{c|cc|cc}
\hline \hline
\specialrule{0em}{2pt}{1pt}
   & \multicolumn{2}{c|}{Model-based Solution} & \multicolumn{2}{c}{Learning-based Method$^\dag$} \\ \specialrule{0em}{1pt}{1pt} \cline{2-5} \specialrule{0em}{1pt}{1pt}
   & \makecell[c]{Analytical\\Model~\cite{Rolf12_IROS}} & \makecell[c]{Numerical \\ Model~\cite{Fang20_TRO}} & \makecell[c]{Direct IK \\ Learning~\cite{Giorelli13_IROS}} & This work \\ \specialrule{0em}{1pt}{1pt} \hline  \specialrule{0em}{1pt}{1pt}
IK Time & 12.6 ms & 138 s &  3.2 ms & 28 ms \\
\specialrule{0em}{1pt}{1pt}
Ave. Error &  8.2 mm  & 4.7 mm  & 0.97 mm  & 0.78 mm \\ 
\specialrule{0em}{1pt}{1pt}
\hline \hline
\end{tabular}
\begin{flushleft}
$^\dag$The same dataset was applied to learn the direct IK mapping~\cite{Giorelli13_IROS} and sim-to-real network $\mathcal{N}_{s2r}$ in our pipeline.
\end{flushleft}
\label{tab:IKCompare}
\end{table}

\begin{figure}[t]\centering 
\includegraphics[width=\linewidth]{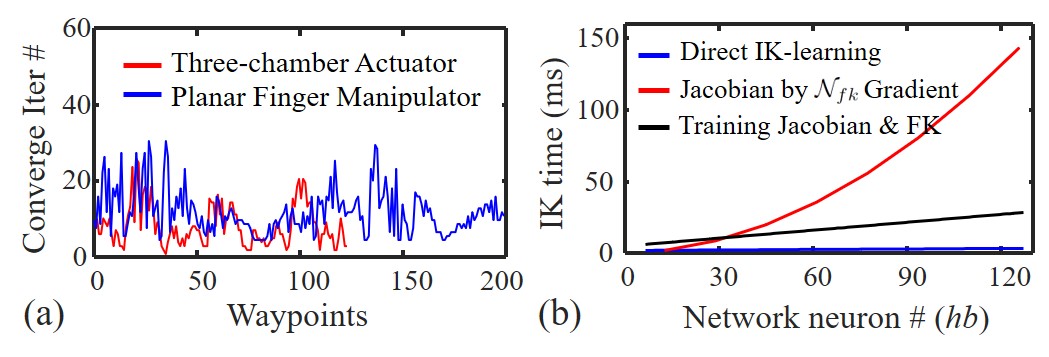}
\caption{Quantitative analysis for (a) the speed of convergence and (b) the time efficiency of our Jacobian-based training method. Three learning strategies for computing IK are compared in (b).}
\label{fig:compSpeed}
\end{figure}

Besides the accuracy, another advantage of our learning-based approach is its low computational cost. Thanks to the efficient forward propagation process of FNN networks and fast converge speed of the Jacobian-based algorithm, the time used to compute single waypoint IK for the three-chamber setup is less than 30ms even when a large number of neurons $hb = 128$ is used. As a final result, the time used to compute IK solutions for the entire `flower' and `box' are $2.53$ sec. and $4.12$ sec., respectively.
The quantitative analysis of the converge speed and IK computing time of our algorithm is showed in Fig.~\ref{fig:compSpeed}, and also compared with other existing solutions (see Table~\ref{tab:IKCompare}). It can be found that learning-based method (for both analytical or numerical methods) can generally provide a more accurate IK result than model-based solution as it can well capture the uncertainties that are hard to calibrate (e.g., material shifting, fabrication error, etc.). When compared between learning-based methods, direct IK learning is the fastest (see Fig.~\ref{fig:compSpeed}(b)). Our Jacobian-based method provides the best accuracy and can also ensure a real-time computing speed (i.e., at the rate of $35+$ waypoints per second). What's more important and also as aforementioned in Sec.~\ref{subsec:relatedWork}, our solution can well handle the redundant soft robots to ensure minimum variation when travelling along the trajectory. This is difficult to be handled by direct IK training.

\subsection{Experiment with soft finger manipulator}
The soft finger manipulator shows in Fig.~\ref{fig:setup}(b) is a redundant system, which has a higher DOFs in its actuation space ($m = 3$) than the task space ($n = 2$). Therefore, an input waypoint can have multiple IK solutions. Both the path following and the interactive positioning tasks are conducted to validate the performance of our learning-based IK solver on the redundant systems. 

\begin{figure}
\centering
\includegraphics[width=0.9\linewidth]{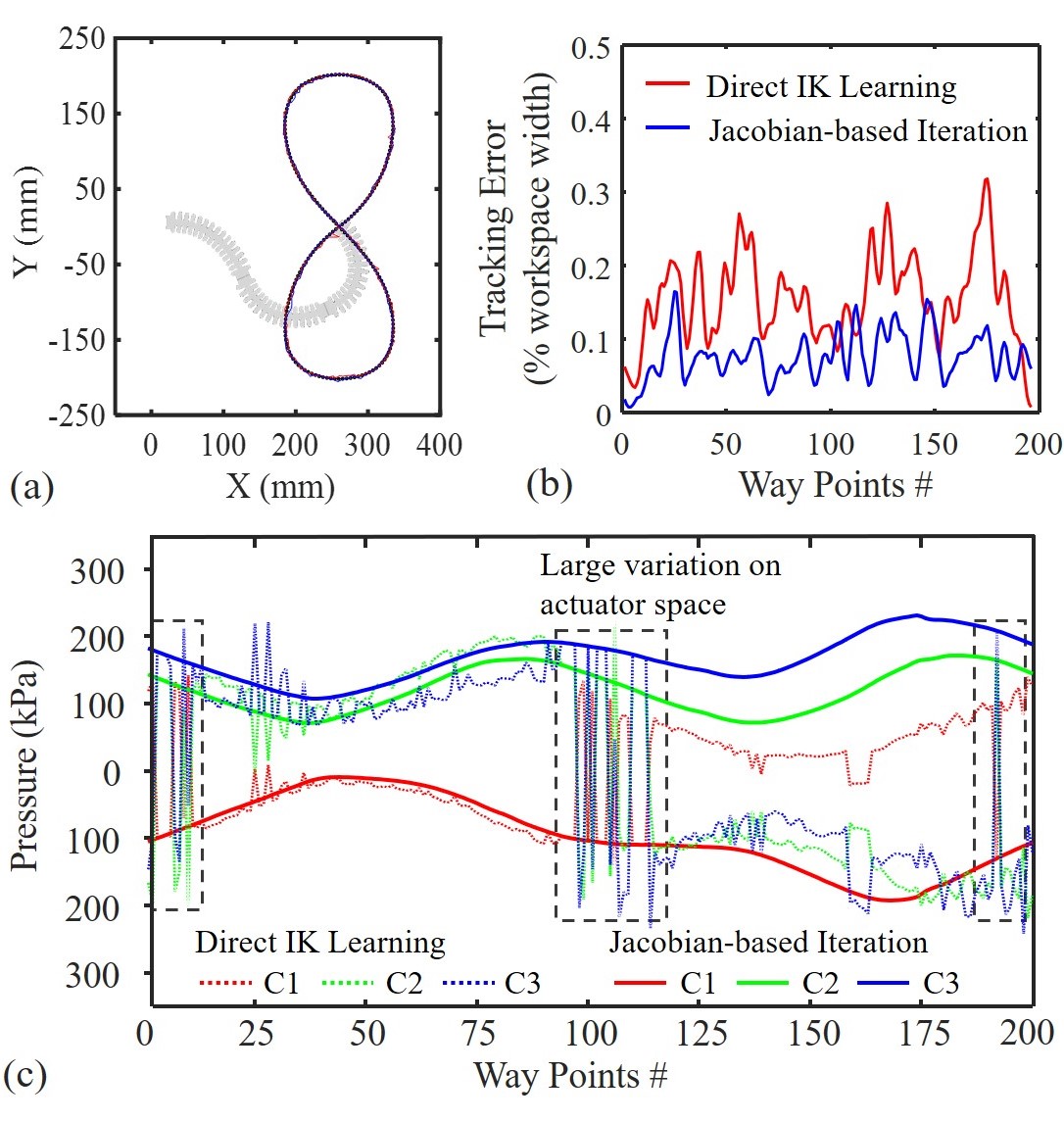}
\caption{(a) Path following result on the soft finger manipulator with the `8'-shape trajectory. (b) Comparison of tracking errors in the task space from the direct IK learning vs. the Jacobian-based iteration by learning (our method). (c) Visualization of IK solutions in the actuator space, where C1, C2 \& C3 present the actuation parameters (i.e., pressures) in three different chambers -- large variation can be found from the results obtained by direct IK learning.} \label{fig:fingerTrajFollowResult}
\end{figure}

\subsubsection{Path following}
We first present the results of following an `8'-shape trajectory that contains 200 way points as shown in Fig.~\ref{fig:fingerTrajFollowResult}(a). The actuation parameters obtained from the Jacobian-based method are compared with those resulting from direct IK-learning. Our Jacobian-based IK by learning demonstrates excellent performance in the accuracy of tracking precision. The average and maximum tracking errors for all waypoints are $0.08\%$ and $0.18\%$ of the workspace width, respectively. As shown in Fig.~\ref{fig:fingerTrajFollowResult}(c), our method is able to ensure a smooth motion that minimizes the variation in actuator space. As a comparison, large variation (i.e., jumps) in the actuator space can be found in the results of direct IK-learning (circled by dash lines in Fig.~\ref{fig:fingerTrajFollowResult}(c)). This problem of direct IK learning is mainly caused by its lack of capability to support the one-to-many IK mapping.
%This also leads to \rev{large jump in the configurations -- }{large errors of path following between two neighboring waypoints, which} can \rev{}{also} be found in the supplementary video. 
For this trajectory, the IK solutions can be efficiently computed by our method at the average speed of $39\mathrm{ms}$ per waypoint.

\begin{figure}[t]\centering 
\includegraphics[width=\linewidth]{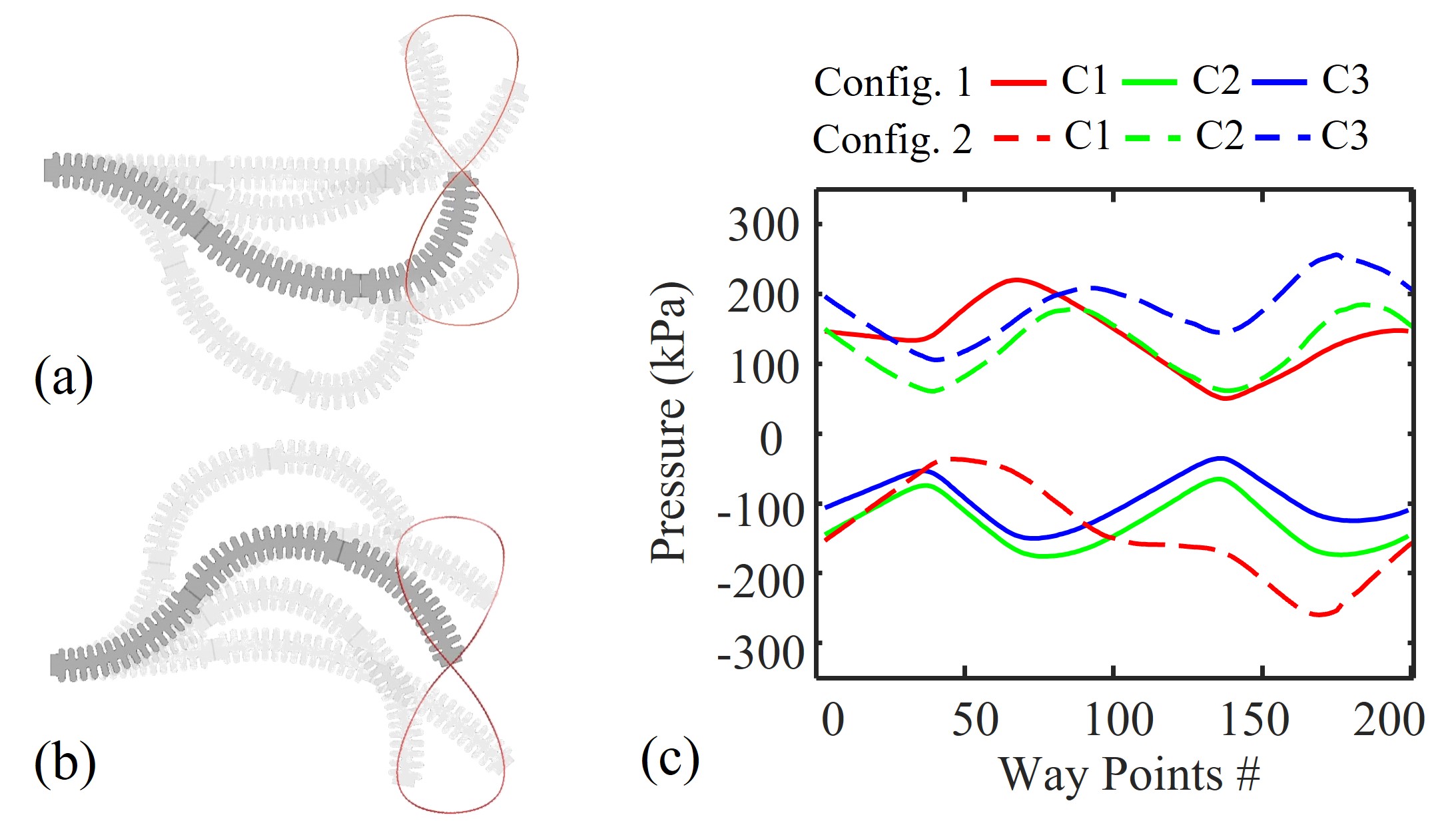}
\caption{Two different configurations of motion are shown in (a) and (b), both of which are feasible IK solutions for following the `8'-shape path by the soft finger manipulator. The smoothness in motion is guaranteed by the Jacobian-based iteration for both results as shown in (c), where the actuation in every chamber has minimal variation in control parameters between neighboring waypoints.}
\label{fig:redoundacyResult}
\end{figure}

\begin{figure}[t]\centering 
\includegraphics[width=\linewidth]{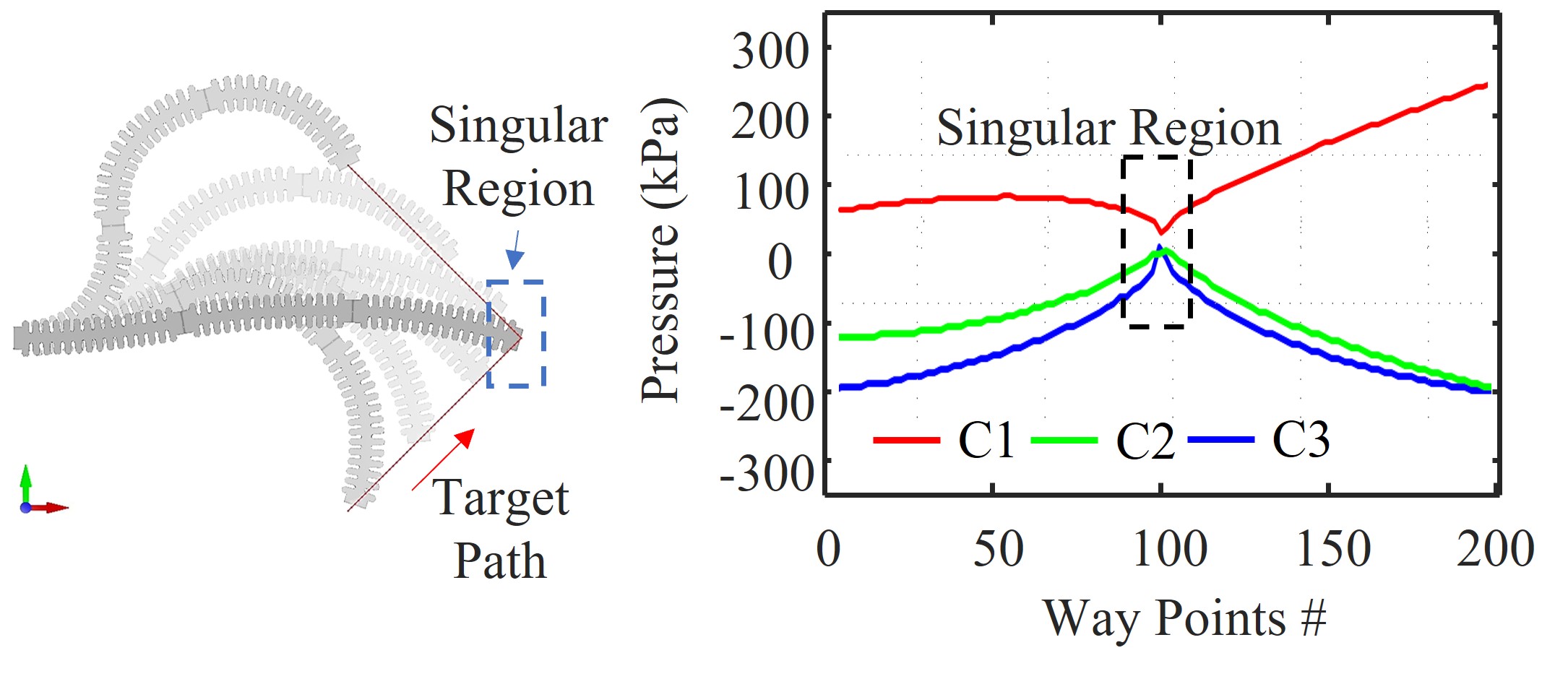}
\caption{Results for the path following task passing through the singularity region. Our method can successfully compute feasible smooth motions when meeting singularity.}
\label{fig:singularityIssue}
\end{figure}

It is worth to mention that the configuration of motion computed by our method is highly dependent on the selection of the IK solution at the starting waypoint (i.e., the initial value). As demonstrated in Fig.~\ref{fig:redoundacyResult}, the configuration of the finger manipulator determined by our IK solver at a waypoint is always close to the IK solution of the previous waypoint where this dependency can trace back to the beginning point. This is because our Jacobian-based iteration tends to minimize the distance-based objective function defined in (\ref{eq:objectives}) while minimizing the change in actuation parameters. This preferred property is also kept when computing IK solutions for a motion passing through the singularity region (see the example in Fig.~\ref{fig:singularityIssue} for following a `L' shape path). As a Jacobian-based iterative solver, our method can always generate a nearly optimal solution for singularity points by applying appropriate terminal conditions (e.g., minimal variation in the value of the objective function). 

\begin{figure} [t]
\centering
\includegraphics[width=0.9\linewidth]{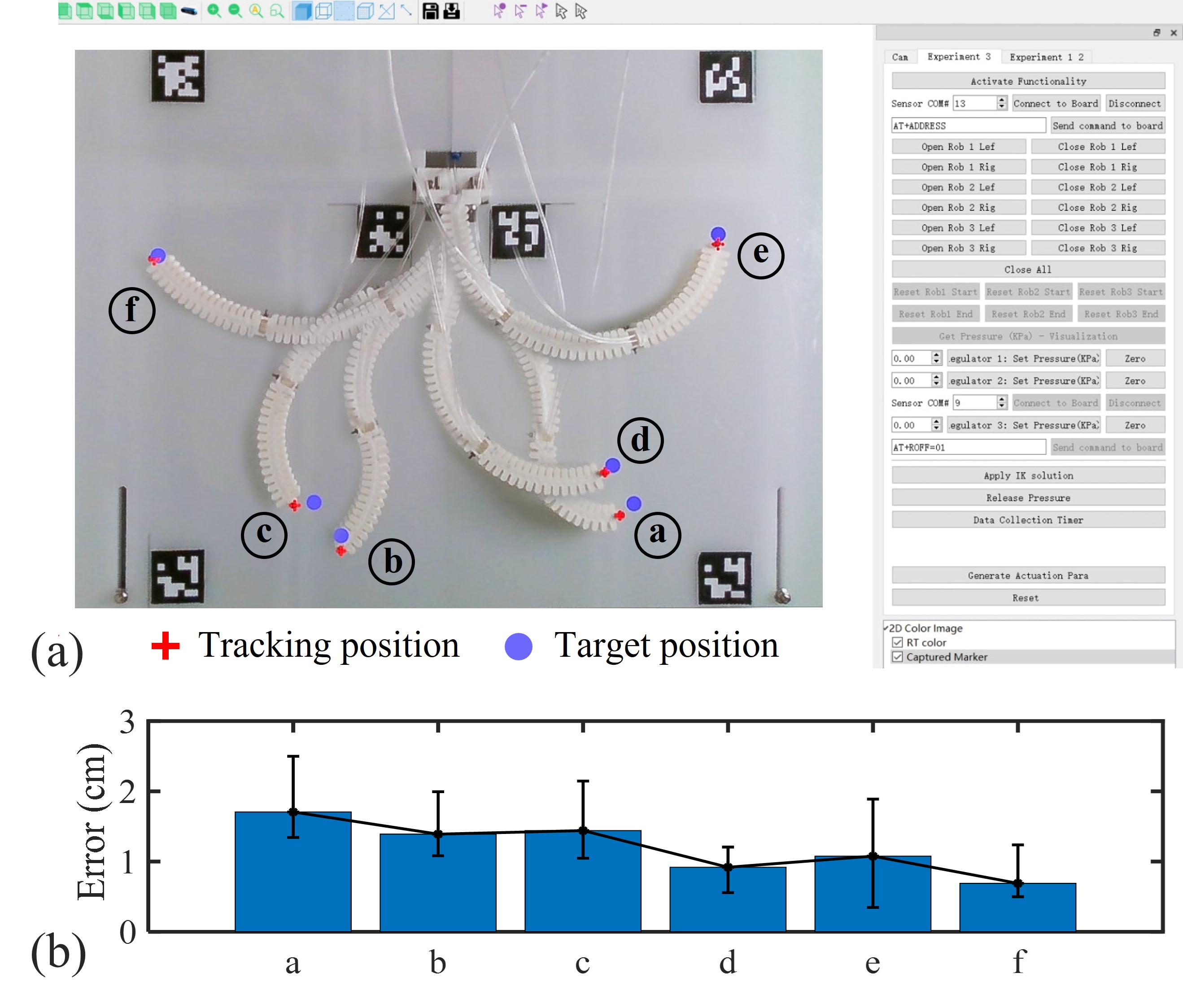}
\caption{Interactive positioning results for the soft manipulator with three finger actuators. (a) With a user-specified position given through the software interface, our Jacobian-based method is applied to determine the IK solution. (b) The bar chat presents the tracking errors for different target positions, where the repeatability is also studied and displayed as the range of deviation in tracking errors.}\label{fig:fingerInteractiveDrag}
\end{figure}
      
\subsubsection{Interactive positioning} \label{subsec:positioning}
The experiment of interactive positioning is also conducted on the soft finger setup. As shown in Fig.~\ref{fig:fingerInteractiveDrag}(a), users can select the desired point location for the manipulator's tip through our interface and our planner will compute the IK solutions as the corresponding actuation parameters. The computation can be efficiently completed at an average speed of 47ms together with the sim-to-real network $\mathcal{N}_{s2r}$. As a result, users can interactively position the manipulator's tip -- see also the supplementary video. When different positions are selected in the work space, the soft manipulator can move among configurations with large variations. 
The errors of positioning are evaluated and presented in Fig.~\ref{fig:fingerInteractiveDrag}(b) as a bar chart. It is found that all six target positions can be realized in the physical environment with tracking errors less than $0.9 \%$ of the work space's width. Note that, each of these six target positions is tested $10$ times in random order to study the repeatability of our system. The results are displayed as the range of derivation on the bar chart. 
   
It is observed that our method can generate results in different configurations for two close waypoints (e.g., the point b and c shown in Fig.~\ref{fig:fingerInteractiveDrag}) when using initial values that are always far from the resultant configurations (zero is adopted for all actuation parameters in this case). Together with the results presented in trajectory following experiments (e.g., Fig.~\ref{fig:redoundacyResult}), our method shows the capability of determining one `nearest' IK solution among all feasible IK solutions.

\subsection{Statistical analysis for sim-to-real transfer}\label{subsec:discussion}

\begin{figure}[t]\centering 
\includegraphics[width=\linewidth]{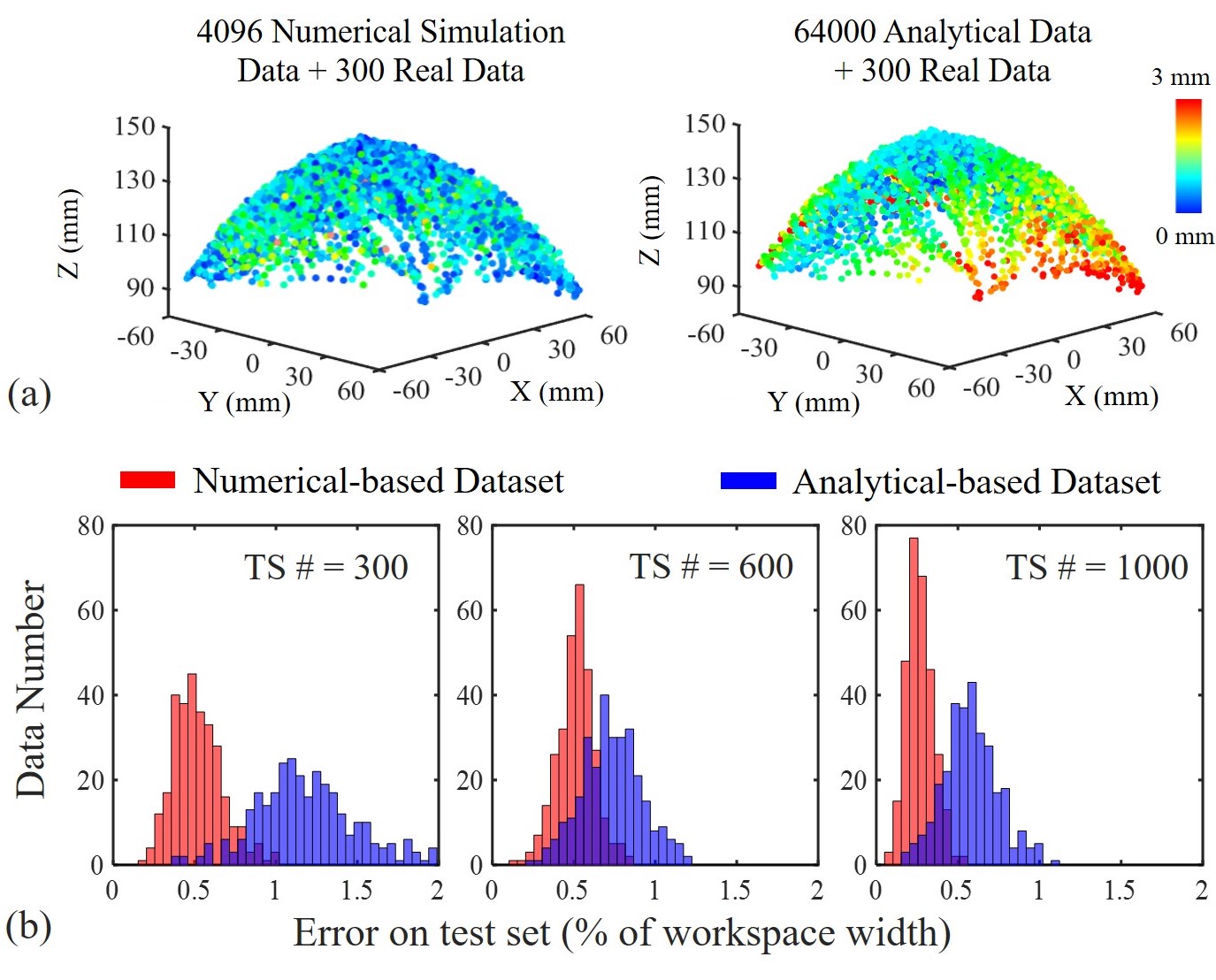}
\caption{Comparison for studying the performance of sim-to-real learning on 1) numerical simulation-based (our approach) vs. 2) analytical computation-based (ref.~\cite{Rolf12_IROS, Drotman18_TMECH}) training datasets. (a) Visualization of prediction errors in the task space. (b) Histograms show the distributions of prediction errors when using different numbers (mark as TS \#) of samples to train $\mathcal{N}_{s2r}$.}
\label{fig:anlyticalCompare}
\end{figure}

Experiments have been conducted on the setup of three-chamber soft robot to explicitly compare the behavior of sim-to-real transfer by using different models in the virtual environment. 
Compared to the numerical simulator used in this work, the reality gap becomes larger when the dataset for training $\mathcal{N}_{fk}$ and $\mathcal{N}_{J}$ is obtained from an analytical model (ref.~\cite{Rolf12_IROS, Drotman18_TMECH}). 
%\rev{}{which has less accuracy to capture the physical behavior of soft robot (this can also be seen with the comparison in Table~\ref{tab:IKCompare})} 
When using the analytical model, more samples are needed for training the sim-to-real network $\mathcal{N}_{s2r}$ to achieve the similar accuracy.
As observed in Fig.~\ref{fig:anlyticalCompare}(a), 
the model trained by the dataset obtained from the numerical simulation (i.e., our approach) shows smaller prediction errors when using the same number of samples to learn the sim-to-real network $\mathcal{N}_{s2r}$. Note that this less accurate result is still observed even after using more samples generated from the analytical model (e.g., 64000 in our experiment) when the number of real samples is fixed.

This experiment also proves that the sim-to-real network can effectively eliminate the gap between simulation and reality -- although requiring different numbers of samples for the model learned from numerical simulation (i.e., our method) and the analytical model. Fig.~\ref{fig:anlyticalCompare}(b) shows that the accuracy within 1\% of the workspace width can be obtained at most points when TS\# = 300 and TS\# = 1000 are used by our method and the analytical model respectively.
As a general case, it is more costly to generate a large number of physical training samples than simulation-based samples. Generating a dataset of empirical samples in a large number is very time-consuming and may result in material failure. 
Our method proposed in this paper converts this challenge into an approach that is easier to realize, in other words, learning a more accurate predictor from more accurate samples generated by numerical simulation. As a result, the gap between prediction and reality can be reduced and fixed by a light sim-to-real network.

\begin{figure}[t]\centering 
\includegraphics[width=\linewidth]{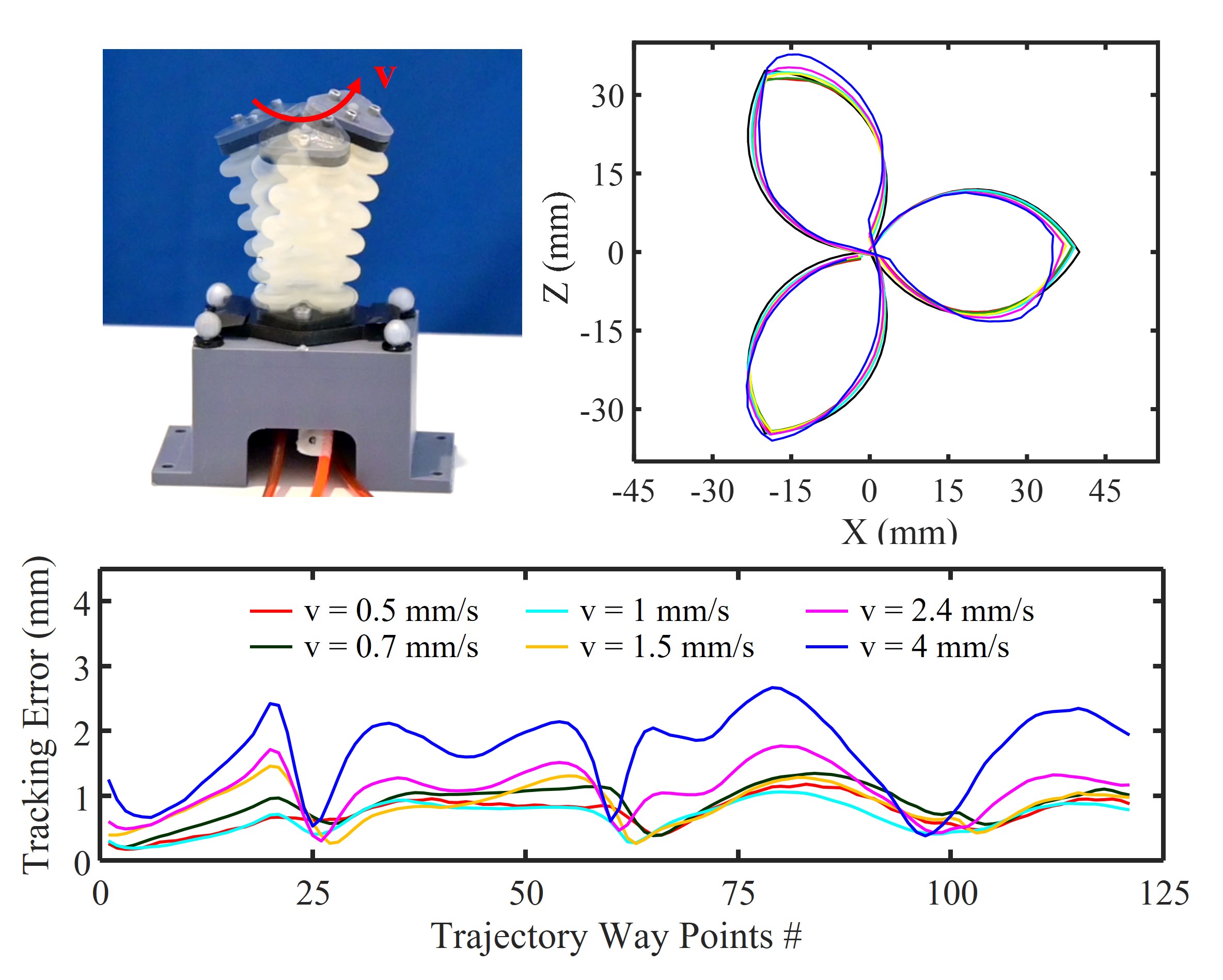}
\caption{Comparison of the behavior in path following by different speeds of motion on the three-chamber soft robot.}
\label{fig:motionSpeed}
\end{figure}

\section{Discussion}

This section discusses the limitation of our learning-based IK solver, where a rigorous analysis is conducted to show the influence with different velocities and external loads on soft robot. On the other hand, the possible extension of pose estimation by our method is presented.

In the proposed Jacobian-based learning, we focus on the computation of quasi-static kinematics. The training datasets in both virtual and physical setups are collected under the quasi-static status, where the hysteresis problem in soft materials is neglected. 
As one major limitation of this work, only considering the quasi-static model will bring large errors in the task of path-following when the dynamic behavior of soft robot is performed (e.g., the velocity of motion is high).
Experiments are conducted to study the influence of speed in motion to the trajectory's accuracy, and the results are shown in Fig.~\ref{fig:motionSpeed}. When the speed of motion is set as less than 1.5 mm/s, the tracking errors can be controlled less than 1.3mm. However, when the speed is increased to 4 mm/s -- i.e., the maximum speed that actuation system presented in Fig.~\ref{fig:setup}(c) can support -- a relatively large error (i.e., the maximal error as 2.7 mm) can be found on the trajectory.
One possible solution to incorporate the hysteresis property of soft materials and dynamic behavior of soft robots in IK computing is to apply the time-variant network structure (e.g., recurrent neural network~\cite{Thuruthel19_TRO}) based on datasets generated in different velocities. This requires much larger training datasets, the generation of which is very time-consuming~\cite{Li18_TMECH}.

\begin{figure}[t]\centering 
\includegraphics[width=\linewidth]{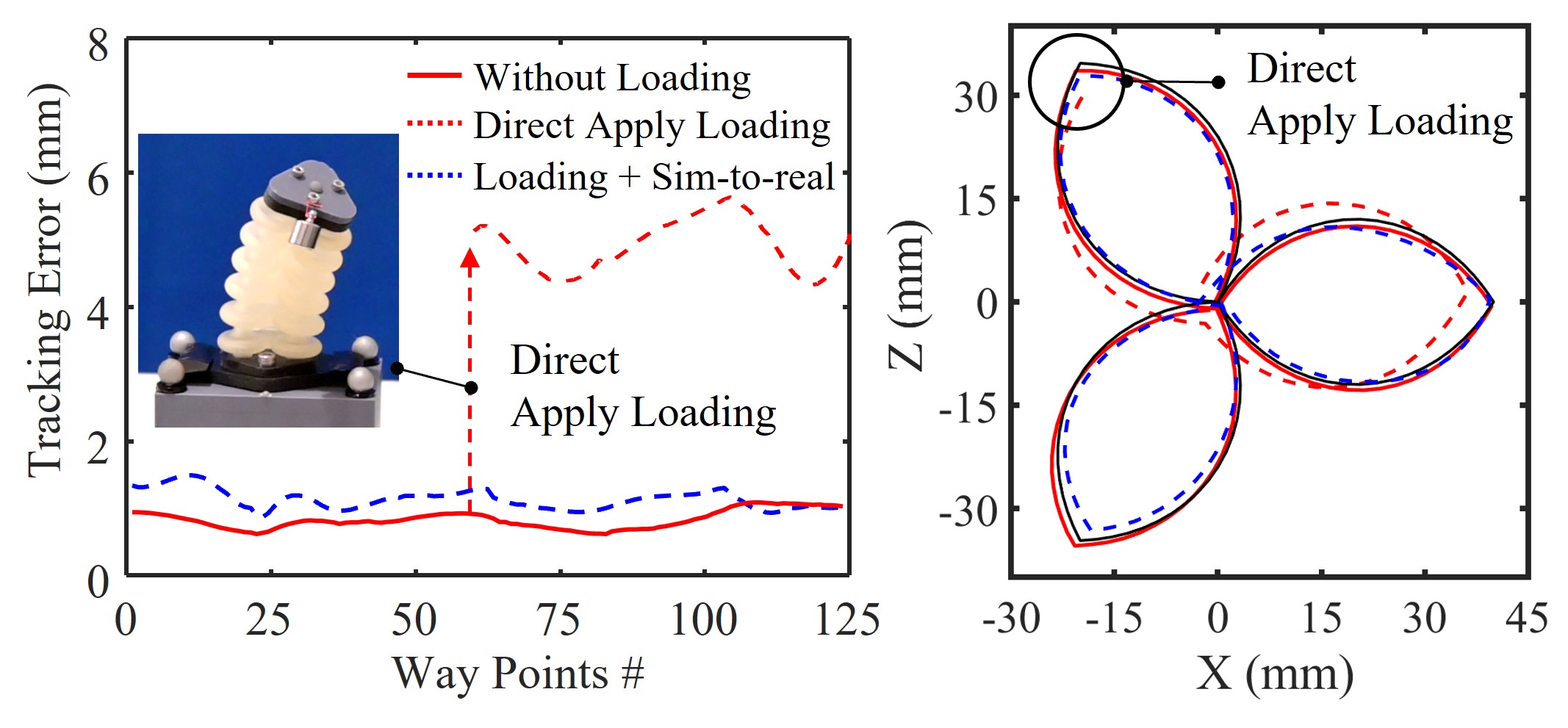}
\caption{Performance of the proposed IK solver when external load is applied to the end-effector, where black shows the target trajectory.}
\label{fig:loadingCase}
\end{figure}

In this work, PID controller is applied to achieve stable pressure to drive soft robots. The vision system is only used to generate ground-truth data for the training purpose. In the verification process, open-loop system is applied to allow soft robots to achieve given tasks. The trained network with sim-to-real transfer can already achieve an acceptable precision in motion without using closed-loop controller.

On the other aspect, the trained sim-to-real network will lose its capability to accurately compute the IK solution when external loads are added to the end-effector of a soft robot. As shown in Fig.~\ref{fig:loadingCase}, significantly enlarged tracking errors can be observed when external load is directly applied (present by dash lines in red). As an additional test, we train a new sim-to-real network by using the soft robot with load. It can be found that the tracking errors become small again by using this newly trained network. This demonstrates the functionality of the sim-to-real transfer to handle external loads.

Another drawback of this work is that we neglect pose information in the pipeline. As an important extension of the learning framework, poses of a soft robot (i.e., including orientation) are to be considered~\cite{Grassmann19_RSS, Peretroukhin20_RSS}. One possible solution is to directly add the rotation into the output layers of $\mathcal{N}_{fk}$ and $\mathcal{N}_{J}$. Meanwhile, training positions and orientations together needs to consider the balance between their different units. Higher DoFs in actuator space are needed to enhance the feasibility of IK solutions.

\section{Conclusion}
In this paper, we present a method to train the forward kinematic model and its Jacobian together as two neural networks to realize the real-time computation of inverse kinematics on soft robots, which is formulated as an optimization problem. As our method can generate smooth motion in a redundant system, it outperforms the existing approaches of direct IK learning. Considering the difficulty in generating large datasets on hardware setups, we adopt a highly effective simulator to generate the training datasets and later apply a sim-to-real network to transfer the kinematic model onto hardware. A lightweight network is employed for sim-to-real mapping so that it can be trained by using a small number of samples. This sim-to-real strategy allows our approach to work on different soft robots that have variations caused by materials and fabrication processes. The main advantages of our method include the efficient computation and the ease of applying the sim-to-real learning transfer.

We test the behavior of our learning-based method in the tasks of path following and interactive positioning on two different soft robotic setups. Our method can solve the IK problem for soft robots effectively and make a good control for the kinematic tasks. As a future work, we plan to explore the possibility of using time-related data for sim-to-real transfer learning that may further enhance the accuracy of IK computing. Moreover, it is also interesting to develop a more transferable learning pipeline that makes the trained model of kinematics can be easily applied to similar designs (e.g., when only the sizes of soft robots are changed).

% \section*{Acknowledgment}
% The authors would like to thank the support from the CUHK Direct Research Grant (CUHK/4055094). Guoxin Fang is partially supported by the China Scholarship Council.

% \clearpage
{\small
\bibliographystyle{IEEEtran}
\bibliography{references}
}

\begin{IEEEbiography}[{\includegraphics[width=1in,height=1.25in,clip,keepaspectratio]{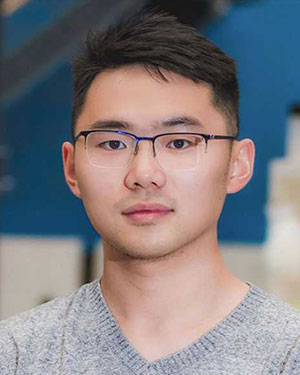}}]{Guoxin Fang}
(Member, IEEE) received the B.E. degree in mechanical engineering from the Beijing Institute of Technology, Beijing, China, in 2016. He is currently working toward the Ph.D. degree in advanced manufacturing with the Delft University of Technology, Delft, The Netherlands.

He is currently also as a Research Assistant with the Department of Mechanical, Aerospace and Civil Engineering, The University of Manchester, Manchester, U.K. His research interests include computational design, digital fabrication, and robotics.
\end{IEEEbiography}

\begin{IEEEbiography}[{\includegraphics[width=1in,height=1.25in,clip,keepaspectratio]{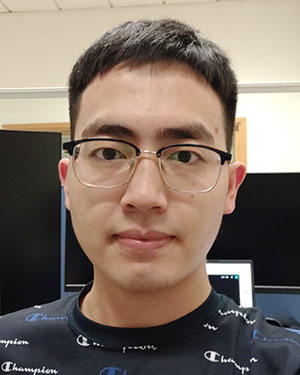}}]{Yingjun Tian} received the bachelor’s degree of engineering in mechanical engineering and automation from the University of Science and Technology of China, Hefei, China, in 2019. He is currently working toward the Ph.D. degree with the Smart Manufacturing Group, Department of Mechanical, Aerospace and Civil Engineering, The University of Manchester, Manchester, U.K. His research interest includes computational design and shape control of soft robots.
\end{IEEEbiography}

\begin{IEEEbiography}[{\includegraphics[width=1in,height=1.25in,clip,keepaspectratio]{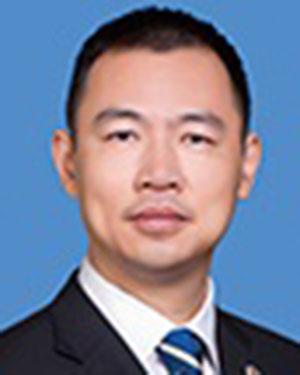}}]{Zhi-Xin Yang}
(Member, IEEE) received the B.Eng. degree in mechanical engineering from the Huazhong University of Science and Technology, Wuhan, China, in 1992, and the Ph.D. degree in industrial engineering and engineering management from the Hong Kong University of Science and Technology, Hong Kong, in 2000. He is currently an Associate Professor in Electromechanical Engineering with the State Key Laboratory of Internet of Things for Smart
City, the Department of Electromechanical Engineering, Faculty of Science and Technology. His current research interests include fault diagnosis and prognosis, machine learning, and computer vision based robotic control.
\end{IEEEbiography}

\begin{IEEEbiography}[{\includegraphics[width=1in,height=1.25in,clip,keepaspectratio]{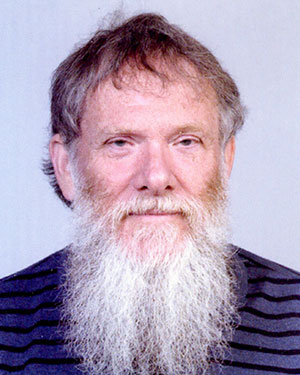}}]{Jo M. P. Geraedts} received the Ph.D. degree in physics from Radboud University, Nijmegen, The Netherlands, in 1983. He then joined Océ, today a Canon group company, where he worked on the development of digital print processes and workflow for document and industrial printing. From 2000 to 2013, he was a Manager with the Océ Industrial Design Department. 
%and was responsible for product, graphic, user interaction, and usability design of all hardware and software developments in multidisciplinary teams worldwide. 
In 2008, he became a Full Professor and the Chair of Mechatronic Design with the Faculty of Industrial Design Engineering, Delft University of Technology, Delft, The Netherlands. His research interests include 3-D scanning, 3-D multimaterial printing, digital reproduction of fine arts, digital manufacturing, and soft robotics.
\end{IEEEbiography}

\begin{IEEEbiography}[{\includegraphics[width=1in,height=1.25in,clip,keepaspectratio]{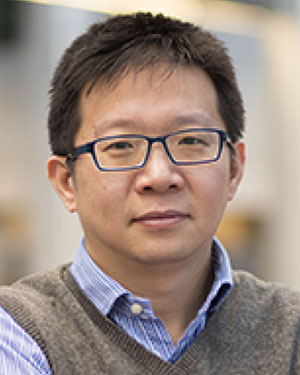}}]{Charlie C.L. Wang} (Senior Member, IEEE) received the B.Eng. degree in mechatronics engineering from the Huazhong University of Science and Technology, China, in 1998, and Ph.D. degrees in mechanical engineering from The Hong Kong University of Science and Technology in 2002. He is currently Professor and Chair of Smart Manufacturing with The University of Manchester, Manchester, U.K. His research interests include digital manufacturing, computational Design, additive Manufacturing, soft Robotics, Mass Personalization, and Geometric Computing. He was elected as a Fellow of the American Society of Mechanical Engineers, in 2013.
\end{IEEEbiography}

\end{document}